\documentclass[journal]{IEEEtran}

\usepackage{amsmath}
\usepackage{amssymb}
\usepackage{booktabs}
\usepackage{caption}
\usepackage{cite}
\usepackage{gensymb}
\usepackage{graphicx}
\usepackage[force,almostfull]{textcomp}
\usepackage{mathtools, cuted}
\usepackage{multirow}
\usepackage{rotating}
\usepackage{subcaption}
\usepackage{tablefootnote}
\usepackage{url}

\newcommand{\etal}{\mbox{\emph{et al.}}}

\begin{document}

\title{Recent Developments in Aerial Robotics: \\
A Survey and Prototypes Overview}


\author{Chun Fui Liew,
        Danielle DeLatte,
        Naoya Takeishi,
        Takehisa Yairi
\thanks{Chun Fui Liew is with the Hongo Aerospace Inc. and graduated from the University of Tokyo, Hongo, Bunkyo-ku, Japan 113-8654 (email: liew@ailab.t.u-tokyo.ac.jp or chun.fui.liew@gmail.com).}
\thanks{Naoya Takeishi, Danielle DeLatte, 
and Takehisa Yairi are with the Department of Aeronautics and Astronautics Engineering, Graduate School of Engineering, University of Tokyo, Hongo, Bunkyo-ku, Japan 113-8654 (email: takeishi@ailab.t.u-tokyo.ac.jp; delatte@ailab.t.u-tokyo.ac.jp; eto@ailab.t.u-tokyo.ac.jp; 	yairi@ailab.t.u-tokyo.ac.jp ).}
}

\maketitle


\begin{abstract}
In recent years, research and development in aerial robotics (i.e., unmanned aerial vehicles, UAVs) has been growing at an unprecedented speed, and there is a need to summarize the background, latest developments, and trends of UAV research. Along with a general overview on the definition, types, categories, and topics of UAV, this work describes a systematic way to identify 1,318 high-quality UAV papers from more than thirty thousand that have been appeared in the top journals and conferences. On top of that, we provide a bird's-eye view of UAV research since 2001 by summarizing various statistical information, such as the year, type, and topic distribution of the UAV papers. We make our survey list public and believe that the list can not only help researchers identify, study, and compare their work, but is also useful for understanding research trends in the field.
From our survey results, we find there are many types of UAV, and to the best of our knowledge, no literature has attempted to summarize all types in one place. With our survey list, we explain the types within our survey and outline the recent progress of each. We believe this summary can enhance readers' understanding on the UAVs and inspire researchers to propose new methods and new applications.
\end{abstract}





\IEEEpeerreviewmaketitle

\section{Introduction}
\label{sec:introduction}

%
%
%
%
\IEEEPARstart{U}NMANNED aerial vehicle (UAV) research and development has been growing rapidly over the past decade. In academia, there are more than 60 UAV papers in IEEE/RSJ International Conference on Intelligent Robots and Systems (IROS) and IEEE International Conference on Robotics and Automation (ICRA) in 2016 alone. In the commercial sector, the annual Aerospace Forecast Report~\cite{uav-survey_faa_report_16} released by the United States Federal Aviation Administration (FAA) estimates that more than seven million UAVs will be purchased by 2020. Another recent report~\cite{uav-survey_pwc_report_16} released by PricewaterhouseCoopers (PwC)---the second largest professional services firm in the world---estimates the global market for applications of UAVs at over \$127 billion in 2020.

Despite rapid growth, there is no survey paper that summarizes the background, latest developments, and trends of the UAV research. And, because the number of UAV papers has grown rapidly in recent years, researchers often find it hard to cite all related papers in a short paper. To date, there is no survey that attempt to list all the UAV papers in a systematic way. In this work, we systematically identify 1,318 UAV papers that appear in the top robotic journals/conferences since 2001. The identification process includes screening paper abstracts with a program script and eliminating non-UAV papers based on several criteria with meticulous human checks. We categorize the selected UAV papers in several ways, e.g., with regard to UAV types, research topics, onboard camera systems, off-board motion capture system, countries, years, etc. In addition, we provide a high-level view of UAV research since 2001 by summarizing various statistical information, including year, type, and topic distribution. We believe this survey list not only can help researchers to identify, study, and compare their works, but also is useful for understanding the research trends in the field.

From our survey results, we also find that the types of UAV are growing rapidly. There is a urgent need to have an overview on the UAV types and categories to enhance readers' understanding and to avoid potential confusion. With the UAV papers list, we outline the recent progress of several UAVs for each UAV type that we have surveyed in this work. These include quadcopter, hexacopter, fixed-wing, flapping-wing, ducted-fan, blimp, cyclocopter, spincopter, Coand{\v a}, and various others. To the best of our knowledge, there is no literature that summarizes as many types of UAV in a unified fashion. Along with the UAV figures, we also briefly describe the novelties (either new methods or applications) of each. We believe that the survey results could be a great source of inspiration and continue to push the boundary of UAV research.

The structure of this survey paper is as follows. In Section~\ref{sec:uav-overview}, we first present the definition, types, categories, and topics of UAV research. In Section~\ref{sec:related-works-on-uav-survey}, we review some survey works that are related to UAV research. Then, we explain our survey methodology in Section~\ref{sec:our-survey-methodology} and summarizes our survey results in Section~\ref{sec:survey-results-overview}. In Section~\ref{sec:uav_types}, we outline all types of UAVs that we have surveyed in this work. Lastly, we present our discussion and final remarks in Section~\ref{sec:discussion-and-final-remarks}.

\section{UAV Overview}
\label{sec:uav-overview}

In this section, we define the term UAV formally and introduce a few types of UAVs based on a common classification. Then, we discuss two interesting ideas that have been proposed by UAV researchers to categorize UAVs and summarize the UAV topics briefly with a pie chart.

\subsection{UAV Definition}  
\label{subsec:uav-definition}

Commonly known as a drone, a UAV is an aircraft that can perform flight missions autonomously without a human pilot onboard~\cite{wikipedia_uav, wikipedia_mav} or can be tele-operated by a pilot from a ground station. The UAV's degree of autonomy varies but often has basic autonomy features such as ``self-leveling'' using an inertial measurement unit (IMU), ``position-holding'' using a global navigation satellite system (GNSS) sensor, and ``altitude-holding'' using a barometer or a distance sensor. UAVs with higher degrees of autonomy offer more functions like automatic take-off and landing, path planning, and obstacle avoidance. In general, a UAV can be viewed as a flying robot. In the literature and UAV communities, a UAV also has several other names like micro aerial vehicle (MAV), unmanned aerial system (UAS), vertical take-off and landing aircraft (VTOL), multicopter, rotorcraft, and aerial robot. In this work, we will use the phrases ``UAV'' and ``drone'' interchangeably.

\subsection{UAV Types}  
\label{subsec:uav-types}

\begin{figure}
  \centering
  \includegraphics[trim={0cm 0cm 0cm 0cm},clip,width=0.48\textwidth]{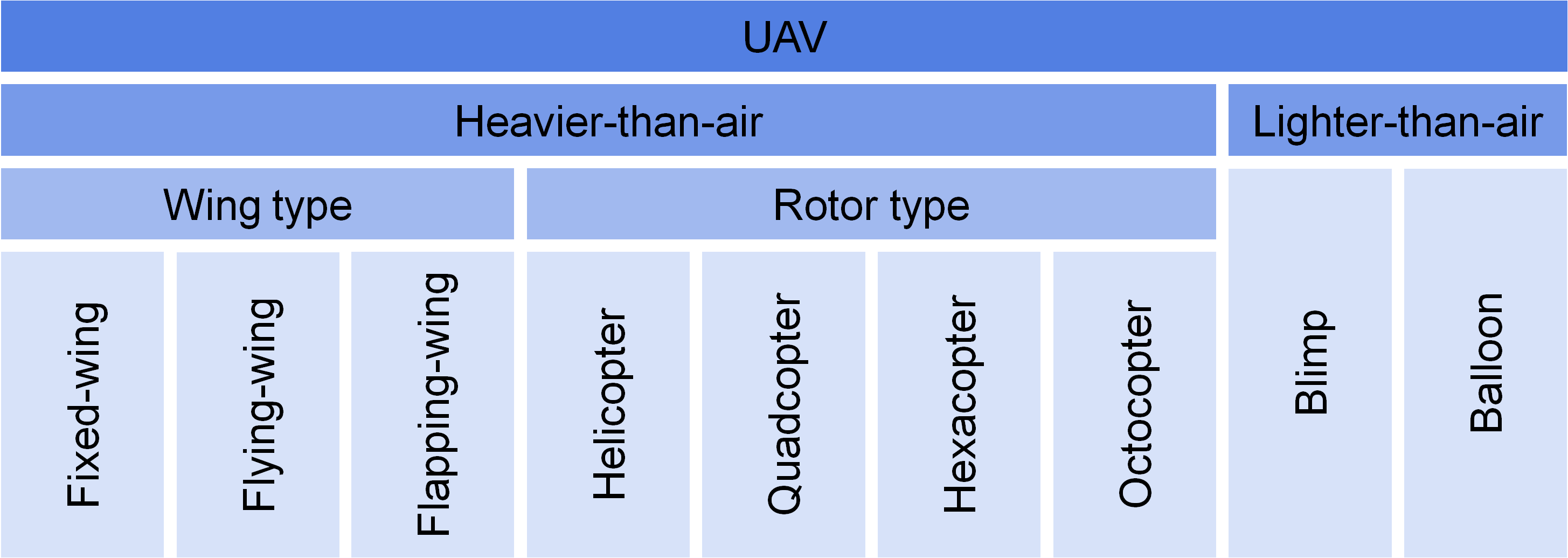}
  \caption{Common UAV types. See text for details.}
  \label{fig:uav-types}
\end{figure}

Depending on the flying principle, UAVs can be classified into several types.\footnote{~Refer to Section~\ref{sec:uav_types} for a comprehensive list of UAV types, along with detailed descriptions and high-resolution figures for each type of UAV.} Figure~\ref{fig:uav-types} illustrates one common classification method, where UAVs are first classified according to their vehicle mass. For example, ``heavier-than-air'' UAVs normally have substantial vehicle mass and rely on aerodynamic or propulsive thrust to fly. On the other hand, ``lighter-than-air'' UAVs like blimps and balloons normally rely on bouyancy force (e.g., using helium gas or heat air) to fly. ``Heavier-than-air'' UAVs can be further classified into ``wing'' or ``rotor'' type. ``Wing'' type UAVs, including fixed-wing, flying-wing, and flapping-wing UAVs, rely on their wings to generate aerodynamic lift; ``rotor'' type UAVs, including a plethora of multirotors, rely on multiple rotors and propellers that are pointing upwards to generate propulsive thrust.

\subsection{UAV Categories}  
\label{subsec:uav-categories}

\begin{figure}
  \centering
  \includegraphics[trim={0cm 0cm 0cm 0cm},clip,width=0.24\textwidth]{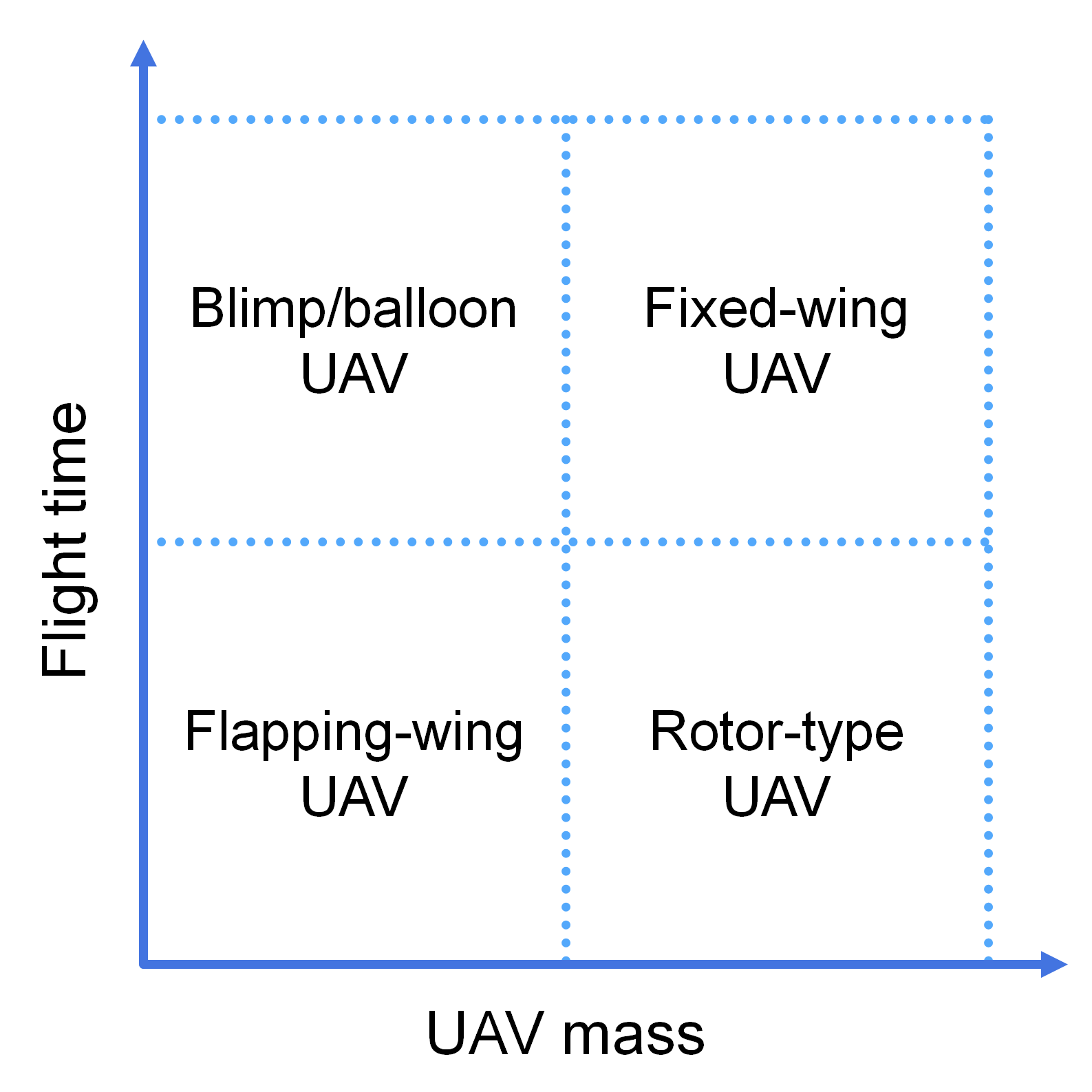}
  \includegraphics[trim={0cm 0cm 0cm 0cm},clip,width=0.24\textwidth]{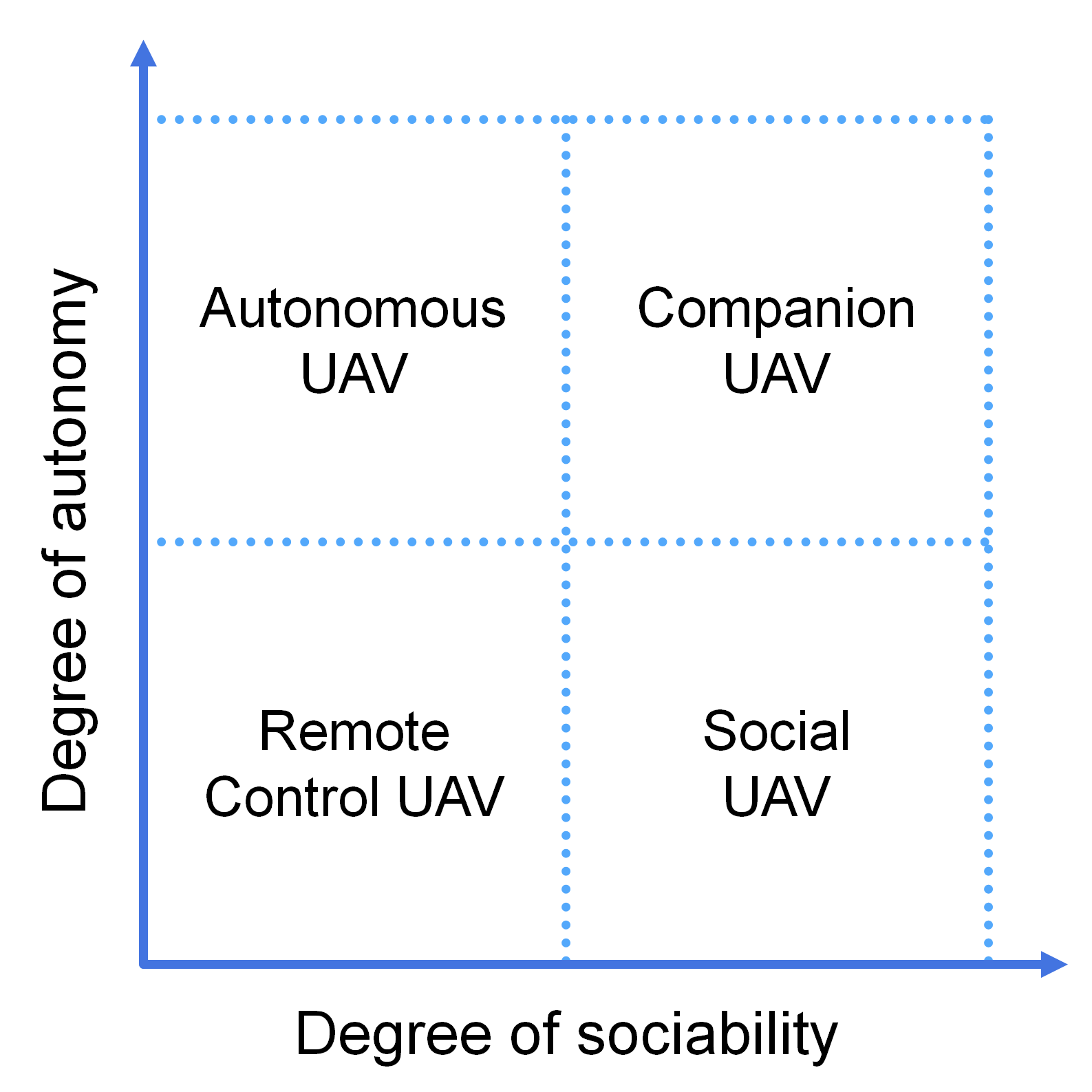}
  \caption{Two UAV categorization methods found in the literature. Left: Flight time versus UAV mass (inspired by Floreano and Wood~\cite{uav_suvey_floreano_15}). Right: Degree of autonomy versus degree of sociability (inspired by Liew~\cite{liew_dissertation_16})). See text for details.}
  \label{fig:uav-categories}
\end{figure}

Previously, we have classified UAVs into several types based on their flying principles. In Fig.~\ref{fig:uav-categories}, Floreano and Wood~\cite{uav_suvey_floreano_15} and Liew~\cite{liew_dissertation_16} provide insights on how UAVs can be categorized with two principal components. 

Floreano and Wood~\cite{uav_suvey_floreano_15} categorize UAVs with two different principal components---flight time versus UAV mass. While they have surveyed 28 different fixed-wing, flapping-wing, and rotor-type UAVs, we simplify their original plot into a conceptual chart in Fig.~\ref{fig:uav-categories} (right). In general, flapping-wing UAVs are usually small and have short flight time. Blimp/balloon UAVs are lightweight and have longer flight time. Rotor-type and fixed-wing UAVs are usually heavier. Assuming the same UAV mass and optimal design, fixed-wing UAVs would have longer flight time than rotor-type UAVs due to their higher aerodynamic efficiency. 

On the other hand, Liew~\cite{liew_dissertation_16} proposes to categorize UAVs based on the degree of autonomy and degree of sociability. Traditionally, UAVs are controlled manually by human operators and have low degrees of autonomy and sociability (remote control UAV). Gradually, along the vertical axis of degree of autonomy, researchers have been improving the autonomy aspects of UAVs, such as better reactive control with more sensors and better path planning algorithms (autonomous UAV). Essentially, autonomous UAVs are less dependent on human operators and are able to perform some simple flight tasks autonomously. On the other hand, along the horizontal axis of degree of sociability, researchers have been improving the social aspects of UAVs, such as designing UAVs that are safe for human-robot interaction (HRI), developing a UAV motion planning model that is more comfortable to humans, and building an intuitive communication interface for UAVs to understand humans (social UAV). Different from autonomous UAVs, social UAVs often have low degree of autonomy. Most HRI researchers solely focus on social aspects and manually control a UAV using Wizard of Oz experiments. Liew~\cite{liew_dissertation_16} first coins the phrase ``companion UAV'', where he defines a companion UAV as one that possesses high degrees of both autonomy and sociability. In addition to the autonomy aspects, such as stabilization control and motion planning, companion UAVs must also focus on the sociability aspects such as safe HRI and intuitive communication interface for HRI. 

\subsection{UAV Topics}  
\label{subsec:uav-topics}

\begin{figure}
  \centering
  \includegraphics[trim={0cm 0cm 0cm 0cm},clip,width=0.48\textwidth]{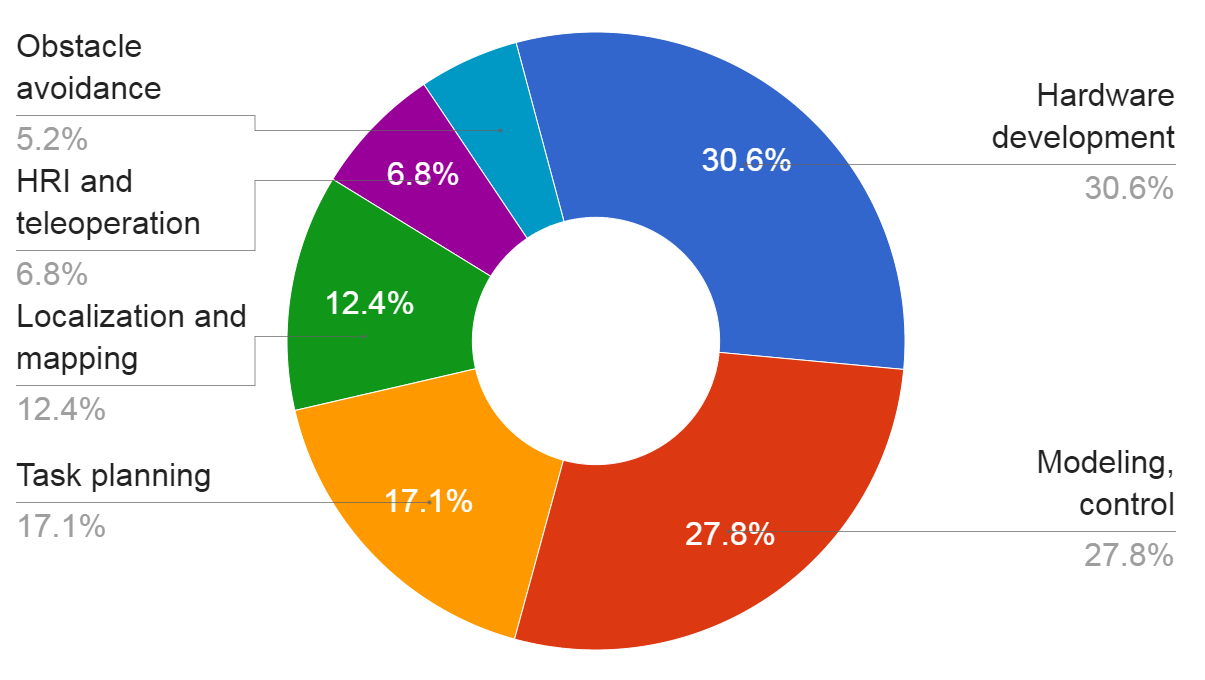}
  \caption{Topic distribution of UAV research in 2006--2016 (data taken from~\cite{liew_dissertation_16}). Best viewed in color. See text for details.}
  \label{fig:uav-topics}
\end{figure}

Focusing on four robotic conferences and four robotic journals, Liew~\cite{liew_dissertation_16} analyzes the topic distribution of UAVs from 2006 to 2016.\footnote{Refer to Section~\ref{sec:our-survey-methodology} \& \ref{sec:survey-results-overview} for a more comprehensive survey and results.} For reference purposes, we simplify the pie chart summarized by Liew~\cite{liew_dissertation_16} in Fig.~\ref{fig:uav-topics}. From the pie chart, we can observe that \textit{hardware} and \textit{control} papers contribute to more than 50\% of the pie. In recent years, researchers start to focus on higher level tasks such as navigation and task planning in UAVs. In addition, researchers also pay attention to visual odometry,  localization, and mapping, which are essential for UAVs to perform task planning effectively. More recently, researchers focus on HRI and tele-operation with UAVs.
Lately, researchers work on obstacle or collision avoidance, which is an important topic of UAVs.

\section{Related Works on UAV Survey}
\label{sec:related-works-on-uav-survey}

In this section, we discuss several surveys in  the UAV field, including surveys on quadcopter and flapping-wing UAVs. We also list several short papers that summarize UAV results in a video. Lastly, we refer to several resources that aim to summarize details of open-source flight controllers.

\subsection{Quadcopter UAV}
\label{subsec:quadcopter-uav}

Focusing on a quadcopter platform, Kumar and Michael~\cite{uav-survey_kumar_ijrr_12} discuss topics on dynamic modeling, trajectory planning, and state estimation in their UAV research. In addition, they outline several challenges and opportunities of formation flight. Different from their work, we consider all types of UAV in this survey paper, including quadcopter, hexacopter, multirotor, fixed-wing, flapping-wing, cyclocopter, coaxial, ducted-fan, glider, blimp, parafoil, kite, Coand{\v a}, and ion-propelled aircrafts (Section~\ref{sec:uav_types}).

\subsection{Flapping-wing UAV}
\label{subsec:flapping-wing-uav}

Wood~\etal~\cite{uav-survey_wood_ijrr_12} present their progress in developing an insect-scale UAV with flapping wings, including topics on dynamic modeling, actuation, control, fabrication, and power. In contrast to their work, we survey the UAV papers since 2001 and provide a general overview of the UAV research, e.g., the number of UAV papers over years (Section~\ref{subsec:yearly-distribution-of-uav-papers}), paper distribution by UAV type (Section~\ref{subsec:uav-types-distribution}), and paper distribution by research topic (Section~\ref{subsec:uav-topics-distribution}), to readers who are interested in this field.

\subsection{Flight Video}
\label{subsec:flight-video}

Ollero and Kondak~\cite{uav-survey_ollero_iros_12} present a video that summarizes the UAV results of four European projects. Similarly, Mellinger~\etal~\cite{uav-survey_mellinger_icra_11} present a video that summarizes some advanced control capabilities of their quadcopter together with a motion capture system, such as flying through a narrow window, robust perching, and cooperative manipulation. On the other hand, Lupashin~\etal~\cite{uav-survey_lupashin_icra_11} present a video that introduces their flying machine arena---an indoor testbed where they use quadcopters and a motion capture system to demonstrate adaptive aggressive flight, iterative learning, rhythmic flight, and balance of an inverted pendulum during flight. 

\subsection{Flight Controller}
\label{subsec:flight-controller}

Lim~\etal~\cite{uav-survey_lim_ram_12} present a survey of the publicly available open-source FCs such as Arducopter, Multiwii, Pixhawk, Aeroquad, OpenPilot, and Paparazzi for UAV. In addition to the hardware details of the FCs, they also discuss the state estimation method and controller structure of each FC. Interestingly, in less than five years, the community has grown very fast and more options are available. Readers who are interested in the latest development of open-source FCs available in the market are recommended to view two recent online articles~\cite{uav-survey_liang_blog_14,uav-survey_liang_blog_15}. 

\section{Our Survey Methodology}
\label{sec:our-survey-methodology}

In this section, we discuss our survey methodology. We first explain the scope of this survey and detail the UAV papers identification process. After that, we describe our on-going plan to update this survey and share the results online.

\subsection{Scope of This Survey}
\label{subsec:scope-of-this-survey}

We cover four top journals and four top conferences in the robotics field since 2001 in this survey. The journals include IEEE Transactions on Robotics (TRO)\footnote{~Known as IEEE Transactions on Robotics and Automation prior to 2004.}, IEEE/ASME Transactions on Mechatronics (TME), The International Journal of Robotics Research (IJRR), and IAS Robotics and Autonomous Systems (RAS); the conferences include IEEE International Conference on Intelligent Robots and Systems (IROS), IEEE International Conference on Robotics and Automation (ICRA), ACM/IEEE International Conference on Human-Robot Interaction (HRI), and IEEE International Workshop on Robot and Human Communication (ROMAN).

\subsection{UAV Papers Identification}
\label{subsec:uav-papers-identification}

The UAV papers identification process involves three major steps. We first use a script to automatically collect more than thirty thousand instances of title and abstract from the mentioned eight journal/conference web pages since 2001, namely TRO, TME, IJRR, RAS, IROS, ICRA, ICUAS, HRI, and ROMAN. We also manually review the hard copies of the IROS and ICRA conferences' table of contents from 2001 to 2004, as we find that not all UAV papers in those years are listed on the website (IEEE Xplore). 

At the second step, we design a list of keywords (Table~\ref{tab:drone-keywords}) to search drone papers systematically from the titles and abstracts collected in the first step. 
Note that we search for both the full name of each keyword (e.g., Unmanned Aerial Vehicle) and its abbreviation (i.e., UAV) with an automatic program script. The keywords include most of the words that describe a UAV. For example, the word ``quadcopter'' or ``quadrotor'' could be detected by the keyword ``copter'' or ``rotor''. As long as one of the keywords is detected, the paper will pass this automated screening process.


\begin{table}[]
  \centering
  \caption{35 keywords used to search drone papers systematically from the collected titles and abstracts.}
  \label{tab:drone-keywords}
  \begin{tabular}{|l|l|l|l|}
    \hline
    acrobatic & bat       & flight   & rotor                                \\
    aerial    & bee       & fly      & rotorcraft                           \\
    aero      & bird      & flying   & soar                                 \\
    aeroplane & blimp     & glide    & soaring                              \\ \cline{4-4}
    air       & copter    & glider   & micro aerial vehicle                 \\
    aircraft  & dragonfly & gliding  & unmmaned aerial vehicle              \\
    airplane  & drone     & hover    & unmanned aircraft system             \\
    airship   & flap      & hovering & vertical takeoff and landing         \\ \cline{4-4}
    balloon   & flapping  & kite     & MAV, UAV, UAS, VTOL                  \\
    \hline
  \end{tabular}
\end{table}

At the third step, we perform a manual screening to reject some non-drone papers. We read the abstract, section titles, related works, and experiment results of all the papers from the second step. If a paper passes all the five criteria below, we consider it a drone paper for this survey.

\begin{enumerate}
  \item The paper must have more than two pages; we do not consider workshop and poster papers.
  \item The paper must have at least one page of flight-related results. These can be either simulation/experiment results, prototyping/fabrication results, or insights/discussion/lesson learned. One exception is a survey/review paper, which normally does not present experiment results. Papers with details/photos of the UAV hardware are a plus. Note that the experiment results do not necessarily need to be a successful flight, e.g., flapping wing UAVs normally have on-the-bench test results.
  \item In topics related to computer vision or image processing, the images must be collected from a UAV's onboard camera rather than a manually moving camera.
  \item In topics related to computer vision or image processing, the images must be collected by the authors themselves. This is important, as authors who collect the dataset themselves often provide insights about their data collection and experiment results.
  \item The paper which proposes a general method, e.g., path planning, must have related works and experiment results on drones. This is important, as some authors mention that their method can be applied to a UAV, but provide no experiment result to verify their statement.
\end{enumerate}

It is interesting to note that using the keyword ``air'' in the second step increases the number of false entries (since the keyword is used in many contexts) but helps to identify some rare drone-related papers that have only the keyword ``air'' in the title and abstract. By manually filtering the list in the third step, we successfully identify two of these drone papers~\cite{uav-background_quadcopter_latscha_iros14, uav-paper_butzke_iros_15}. Similarly, using the keyword ``bee'' can help to identify a rare drone paper~\cite{uav-paper_das_ras_16}. On the other hand, we chose not to use the keyword of ``wing'' because it causes many false entries like the case of ``following'', ``knowing'', etc.

\subsection{Survey Updates and Online Sharing}
\label{subsec:survey-updates-and-online-sharing}

\begin{figure}
  \centering
  \includegraphics[width=0.48\textwidth]{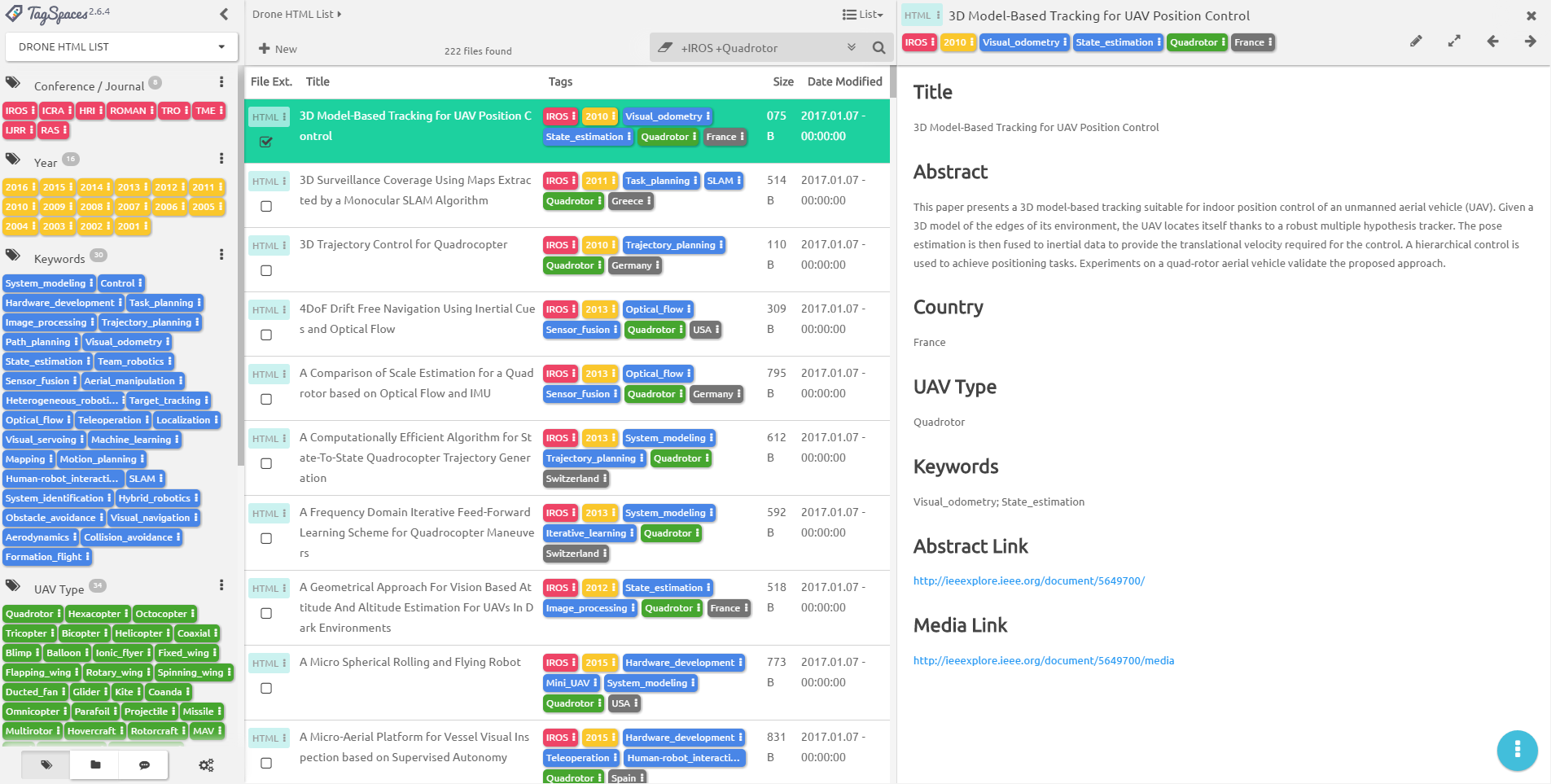}
  \caption{A screenshot of TagSpaces with different categories of tags on the left hand side, list of drone papers that match the search criterion at the middle, and info of the selected paper in HTML format on the right hand side. Best viewed in color.}
  \label{fig:tagspaces-screenshot}
\end{figure}

The full survey results (with all raw information) is shared and updated frequently online via Google Sheets.\footnote{~Tables with full survey results can be viewed on \url{https://goo.gl/cCoCwL}.} 
Major updates, such as additional drone papers from the latest conferences/journals, will be carried out once every three months. While Google Sheets contains all the survey results, we find that it is not possible to tag the papers, and it is also difficult to search multiple keywords in the long paper list effectively. To overcome these issues, we use an open-source file tagging and organization software called TagSpaces~\cite{online_application_tagspaces}. Figure~\ref{fig:tagspaces-screenshot} shows a screenshot of TagSpaces. TagSpaces enables readers to search papers with multiple tags or/and keywords effectively. For example, to search all IROS papers in 2016 that are related to quadcopter, users only need to input ``+IROS +2016 +Quadcopter'' into the search column. Moreover, since original papers (PDF files) cannot be shared with readers due to copyright issues, for each paper entry, we create an HTML file that contains the most important information inside (such as abstract, keywords, country, paper URL link, and video URL link) for easier reference. To setup TagSpaces and download all the HTML files, refer to our website at \url{https://sites.google.com/view/drone-survey}.

\section{Survey Results Overview}
\label{sec:survey-results-overview}

In this section, we give an overview of the survey results, including the year, UAV type, and topic distribution of the UAV papers. For more results, please refer to Appendix~\ref{apx:additional_survey_results}.

\subsection{Yearly Distribution of UAV Papers}
\label{subsec:yearly-distribution-of-uav-papers}

\begin{figure}
  \centering
  \includegraphics[trim={0.5cm 0cm -0.5cm 0cm},clip,width=0.48\textwidth]{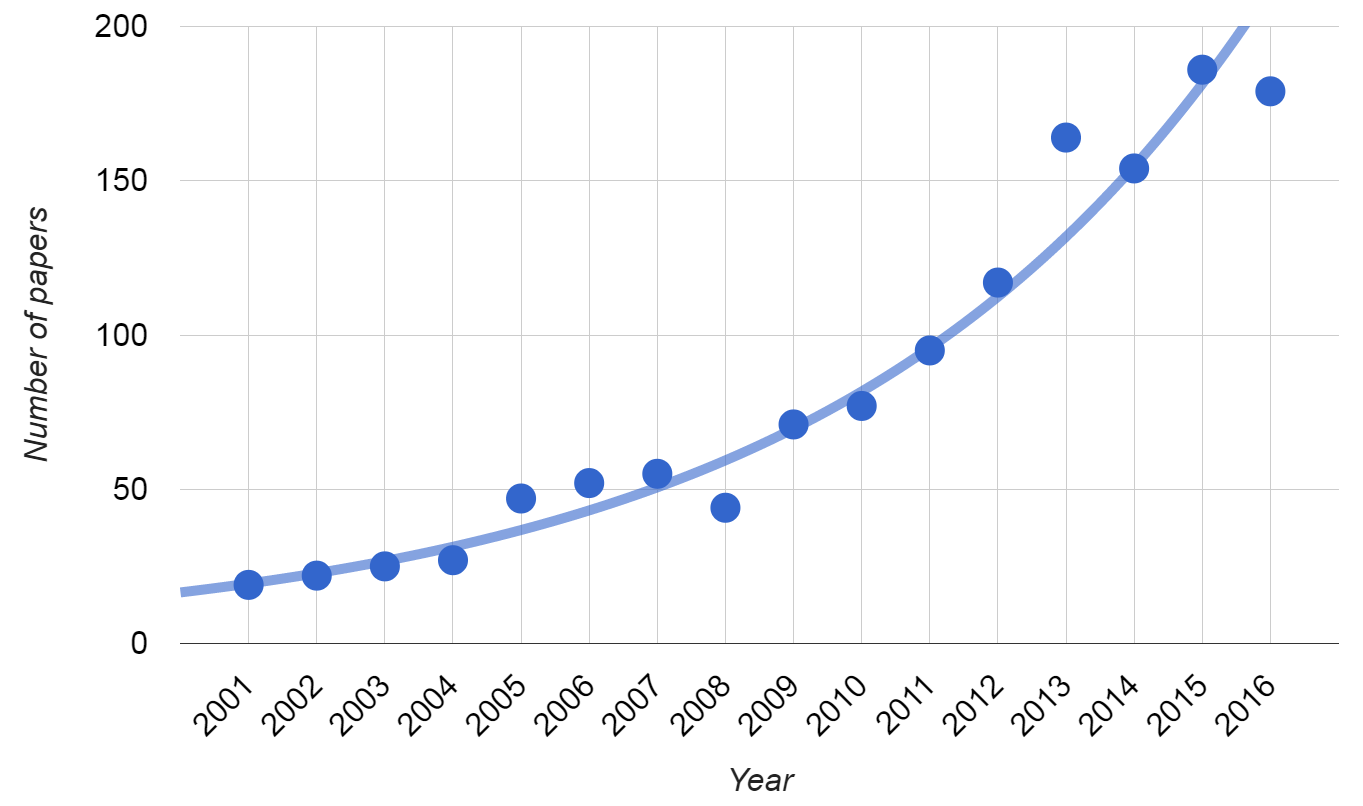}
  \caption{Numbers of UAV papers (dots) identified from the top eight journals/conferences over the years 2001--2016, with an exponential curve fitting plot.}
  \label{fig:number-of-uav-papers}
\end{figure}

Figure~\ref{fig:number-of-uav-papers} plots the numbers of UAV papers identified from the top eight journals and conferences from 2001 to 2016. From the figure, we can observe that the number of UAV papers increases rapidly over the years. As mentioned in the introduction section, the rapid increase is supported by a few factors, such as easier control of the quadcopter configuration, and lower cost of processors and sensors. While there is a slight drop in the number of papers in 2016, with the current strong trends in the research, commercial, government, and hobbyist sectors, we believe that the number of drone papers within the next five years would continue to exceed 150 papers per year.

\subsection{UAV Types Distribution}
\label{subsec:uav-types-distribution}

\begin{table*}
  \centering
  \caption{Top twenty-five keywords from the identified drone papers.}
  \label{tab:top-twenty-five-keywords}
  \begin{tabular}{ll|ll|ll|ll|ll}
    System modeling & 315 & Trajectory generation & 80 & Sensor fusion & 51 & Teleoperation & 45 & Motion planning & 35 \\
    Control & 288 & Path planning & 78 & Aerial manipulation & 47 & Localization & 44 & Human-robot interaction & 33 \\
    Hardware development & 265 & Visual odometry & 75 & Heterogeneous robotics & 47 & Visual servoing & 44 & SLAM & 32 \\
    Task planning & 161 & State estimation & 74 & Target tracking & 46 & Machine learning & 41 & System identification & 32 \\
    Image processing & 117 & Team robotics & 68 & Optical flow & 45 & Mapping & 40 & Hybrid robotics & 31
\end{tabular}
\end{table*}

\begin{figure*}
  \centering
  \includegraphics[trim={0cm 0cm 0cm 0cm},clip,width=1.00\textwidth]{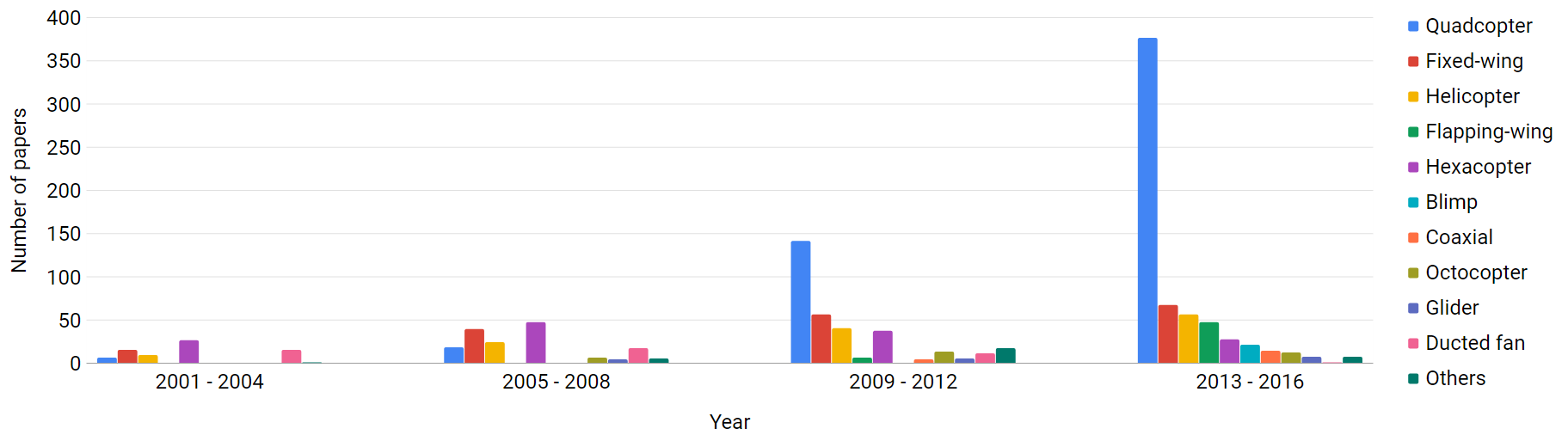}
  \caption{The change of papers distribution by UAV types over the years (overview). Best viewed in color. See text for details.}
  \label{fig:paper-distribution-by-uav-types-overview}
\end{figure*}


Figure~\ref{fig:paper-distribution-by-uav-types-overview} 
shows the UAV types distribution of surveyed papers over different years. The most notable transition in the bar graphs is the number of quadcopter papers, where it increases from 7, 19, 142, to 377 over the past sixteen years. The number of fixed-wing and flapping-wing papers more gradually increases over the years. 

The number of hexacopter papers is zero before 2008. In 2009--2012, it increases to 7; in 2013-2016, it further increases to 45. The number of octocopter papers has a similar pattern. It has zero entries before 2012 but in 2013-2016, the number sudden increases to 222. We believe that hexacopter and octocopter are gaining more attention from researchers, since it has several advantages over quadcopters. First, they have redundant actuation; they are still able to fly/land safely when one motor is malfunctioning without a complex control algorithm. Second, they can handle higher payloads and researchers can mount heavier hardware, such as a robotic arms for an aerial manipulation application, or a 3D lidar sensor for a mapping application. Third, with small modifications, they can perform holonomic flight (move horizontally without tilting motion), where the the UAV is able to achieve 3D force motion without complex coupled dynamic effect, is more robust against wind disturbance and is able to achieve higher flight precision at the same time. 

On the other hand, we notice that since 2005-2008, the number of helicopter papers starts to drop gradually from 48, 38, to 28. The possible cause for this decrease is the difficulties of helicopter control (when compared to quadcopter). Interestingly, the variety of UAVs also increases substantially since 2001. In 2016, in addition to the major six types of UAV (quadcopter, fixed-wing, helicopter, flapping-wing, hexacopter, and blimp), UAV papers also involve topics on coaxial, octocopter, glider, ducted fan, tricopter, bicopter, balloon
, ionic flyer, cyclocopter
, spincopter
, kite, Coand{\v a}, omnicopter
, parafoil, projectile, and missile UAV.

\subsection{UAV Topics Distribution}
\label{subsec:uav-topics-distribution}

Table~\ref{tab:top-twenty-five-keywords} summarizes the top twenty-five keywords in the surveyed UAV papers. From Table~\ref{tab:top-twenty-five-keywords}, we note that \textit{system modeling} and \textit{control} papers are the most frequent keywords. This is not surprising, as most UAVs require system modeling for dynamic control. It has been shown that a simple model-free PID controller is good enough for the basic maneuvers of a UAV~\cite{uav_quadcopter_grzonka_12, uav-survey_lim_ram_12, uav_quadcopter_kawasaki_13, uav_coaxial_bouabdallah_06, uav-background_fixed-wing_bapst_iros15, uav_blimp_burri_13}. For aggressive maneuvers~\cite{uav_control_mellinger_12, uav_control_hehn_13} or more complex dynamics with onboard manipulators~\cite{uav_control_cano_15, uav_control_heredia_14}, dynamic models are normally employed. With a precision indoor positioning system, current state-of-the-art methods have successfully demonstrated formation flights~\cite{uav_control_turpin_12}, flying inverted pendulum~\cite{uav_control_hehn_11}, pole acrobatics~\cite{uav_control_brescianini_13}, ball juggling~\cite{uav_control_dong_15}, cooperative operation~\cite{uav_control_ritz_12, uav_control_nguyen_15, uav_control_ritz_13}, and failure recovery~\cite{uav_control_faessler_15}.

From Table~\ref{tab:top-twenty-five-keywords}, we can also observe that there are large amount of \textit{hardware development} papers since 2001, including papers on quadcopters~\cite{uav_quadcopter_grzonka_12, uav-survey_lim_ram_12, uav-background_quadcopter_kalantari_icra13, uav_quadcopter_kawasaki_13, uav_quadcopter_mizutani_15}, hexacopters~\cite{uav_hexacopter_liew_15-1, uav_hexacopter_liew_15-2, uav_hexacopter_higuchi_12, raabe_dissertation_13}, octocopters~\cite{6696805, uav_octocopter_salazar_09, uav_octocopter_romero_09}, coaxial helicopters~\cite{uav_coaxial_klaptocz_10, uav-background_coaxial_briod_iros13, uav-background_coaxial_paulos_icra15, uav_coaxial_paulos_13, uav_coaxial_bouabdallah_06, uav_coaxial_wang_06}, a helicopter~\cite{uav_helicopter_marantos_14}, a tandem helicopter~\cite{uav_tandem_oh_05}, a bicopter~\cite{uav_bicopter_kawasaki_15}, 
a trirotor UAV~\cite{uav-background_fixed-wing_papachristos_icra13}, fixed-wing UAV~\cite{uav-background_fixed-wing_daler_iros15, uav-background_fixed-wing_bapst_iros15}, flapping-wing UAV~\cite{uav_flapping-wing_hines_15, uav_flapping-wing_park_12, uav_flapping-wing_chin_14, uav_flapping-wing_hamamoto_15, uav-background_flapping-wing_peterson_iros11}, cyclocopter~\cite{uav-background_fixed-wing_pounds_icra13, uav-background_spinning-wing_orsag_icra11}, and blimp UAVs~\cite{uav_blimp_burri_13, uav_blimp_yang_12}.

In recent years, researchers focus on higher level tasks such as navigation and task planning in UAVs~\cite{uav_navigation_dryanovski_13, 6696922, uav_navigation_engel_12}. In addition, researchers also pay attention to visual odometry~\cite{uav_visual-odometry_warren_16, uav_visual-odometry_bloesch_15}, localization~\cite{doi:10.1177/0278364914561101, 5980317, 7139931, uav_localization_konomura_14}, and mapping~\cite{uav_mapping_cole_10, uav_mapping_holz_15, 7353622, 7139761} applications of UAVs. More recently, researchers work on obstacle or collision avoidance~\cite{uav_obstacle-avoidance_bry_15, uav_obstacle-avoidance_ross_13, uav_obstacle-avoidance_roelofsen_15, uav_obstacle-avoidance_muller_14}, which are important topics for UAVs. Current state-of-the-art UAVs could perform robust image-based six Degrees-of-Freedom (DoF) localization~\cite{doi:10.1177/0278364914561101}, cooperate mapping~\cite{uav_mapping_cole_10}, aggressive flight in dense indoor environment~\cite{uav_obstacle-avoidance_bry_15}, and flying through a forest autonomously~\cite{uav_obstacle-avoidance_ross_13}.

More recently, researchers also focus on HRI and tele-operation of UAVs, including jogging UAVs~\cite{uav_hri-platform_graether_12, uav_hri-platform_mueller_15}, a flying humanoid robot~\cite{uav_hri-platform_cooney_12}, a hand-sized hovering ball~\cite{uav_hri-platform_nitta_14}, and various human-following UAVs~\cite{uav_hri-human-tracking_pestana_13, uav_hri-human-tracking_lim_15, uav_hri-human-tracking_naseer_13}.

\section{UAV Types}
\label{sec:uav_types}

In Section~\ref{subsec:uav-types}, we discuss the UAV type distribution within the surveyed papers. To enhance understanding and avoid confusion, in this section, we summarize all types of UAVs that have been proposed by researchers. Along with figures, we briefly describe the novelty of each example UAV, including quadcopter, hexacopter, fixed-wing, flapping-wing, single-rotor, coaxial, ducted-fan, octocopter, glider, blimp, ionic flyer, cyclocopter, spincopter, Coand{\v a}, parafoil, and kite UAVs.

\subsection{Quadcopters}
\label{subsection:state-of-the-art-quadcopters}

\begin{figure*}
  \centering
  \includegraphics[width=\textwidth]{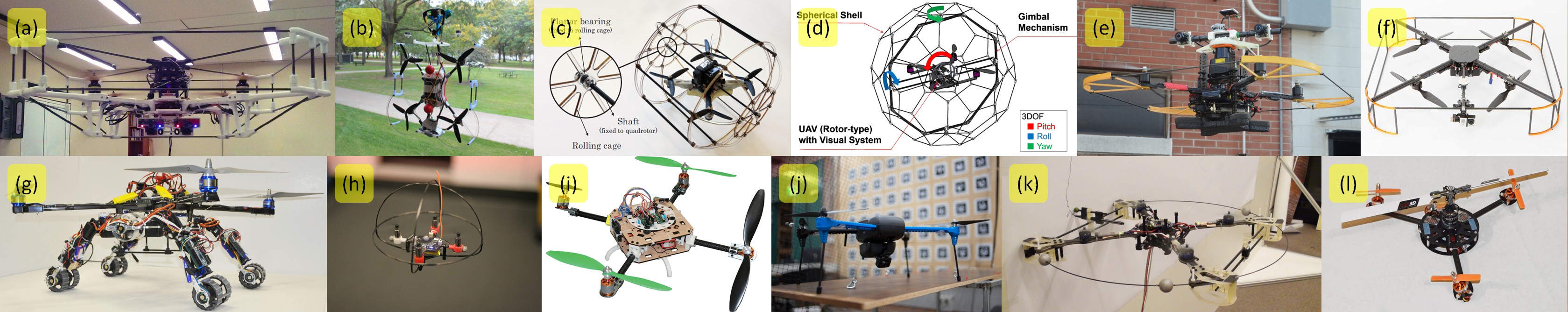}
  \caption{Prototypes of quadcopters (in alphabetical order:~\cite{uav-background_quadcopter_papachristos_iros15, uav-background_quadcopter_kalantari_icra15, uav-background_quadcopter_kalantari_icra13, uav-background_quadcopter_okada_iros16, uav-background_quadcopter_shen_icra14, uav-background_quadcopter_ishiki_iros15, uav-background_quadcopter_latscha_iros14, uav-background_quadcopter_mulgaonkar_icra15, uav-background_quadcopter_oosedo_icra15, uav-background_quadcopter_abeywardena_icra16, uav-background_quadcopter_darivianakis_icra14, uav-background_quadcopter_driessens_iros13}) appear in the reviewed papers. Note that 8 out of the 12 illustrated quadcopters have protective cases for safer operation and human-robot interaction. See text for details.}
  \label{fig:drone-quadcopter-illustration}
\end{figure*}

A quadcopter (Fig.~\ref{fig:drone-quadcopter-illustration}~(a)--(l))~\cite{uav-background_quadcopter_papachristos_iros15, uav-background_quadcopter_kalantari_icra15, uav-background_quadcopter_kalantari_icra13, uav-background_quadcopter_okada_iros16, uav-background_quadcopter_shen_icra14, uav-background_quadcopter_ishiki_iros15, uav-background_quadcopter_latscha_iros14, uav-background_quadcopter_mulgaonkar_icra15, uav-background_quadcopter_oosedo_icra15, uav-background_quadcopter_abeywardena_icra16, uav-background_quadcopter_darivianakis_icra14, uav-background_quadcopter_driessens_iros13} is a UAV with four rotors. Papachristos~\etal~\cite{uav-background_quadcopter_papachristos_iros15} first present an autonomous quadcopter (Fig.~\ref{fig:drone-quadcopter-illustration}~(a)) with do-it-yourself (DIY) stereo perception unit that could track a moving target and perform collision-free navigation. With the goal of reducing quadcopter energy consumption, Kalantari~\etal~\cite{uav-background_quadcopter_kalantari_icra15} build a quadcopter (Fig.~\ref{fig:drone-quadcopter-illustration}~(b)) that uses a novel adhesive gripper to autonomously perch and take-off on smooth vertical walls. Kalantari and Spenko~\cite{uav-background_quadcopter_kalantari_icra13} build one of the first hybrid quadcopters (Fig.~\ref{fig:drone-quadcopter-illustration}~(c)) that is capable of both aerial and ground locomotion. While this quadcopter can only rotate in one direction, Okada~\etal~\cite{uav-background_quadcopter_okada_iros16} present a quadcopter with a gimbal mechanism (Fig.~\ref{fig:drone-quadcopter-illustration}~(d)) that enables the quadcopter to rotate freely in the 3D space. The developed quadcopter is good for inspection applications as the gimbal-like rotating shell helps the quadcopter fly safely in a confined environment with many obstacles.

To the best of our knowledge, Shen~\etal~\cite{uav-background_quadcopter_shen_icra14} is first to present an autonomous quadcopter (Fig.~\ref{fig:drone-quadcopter-illustration}~(e)) that could fly robustly indoors and outdoors by integrating information from a stereo camera, a 2D lidar sensor, an IMU, a magnetometer, a pressure altimeter, and a GPS sensor. Aiming for application in search and rescue missions, Ishiki and Kumon~\cite{uav-background_quadcopter_ishiki_iros15} present a quadcopter (Fig.~\ref{fig:drone-quadcopter-illustration}~(f)) that is equipped with a microphone array to perform sound localization. While the hybrid quadcopter shown in Fig.~\ref{fig:drone-quadcopter-illustration}~(c))~\cite{uav-background_quadcopter_kalantari_icra13} is designed to roll on flat ground, Latscha~\etal~\cite{uav-background_quadcopter_latscha_iros14} combine a quadcopter with two snake-like mobile robots (Fig.~\ref{fig:drone-quadcopter-illustration}~(g)), which make the resulting hybrid robot able to move effectively in disaster scenarios. To increase the safety and robustness of a swarm of quadcopter, Mulgaonkar~\etal~\cite{uav-background_quadcopter_mulgaonkar_icra15} design a small quadcopter with a mass of merely twenty-five grams (Fig.~\ref{fig:drone-quadcopter-illustration}~(h)).

Aiming for higher performance, Oosedo~\etal~\cite{uav-background_quadcopter_oosedo_icra15} develop an unique quadcopter (Fig.~\ref{fig:drone-quadcopter-illustration}~(i)) that could hover stably at various pitch angles with four tiltable propellers. Abeywardena~\etal~\cite{uav-background_quadcopter_abeywardena_icra16} present a quadcopter (Fig.~\ref{fig:drone-quadcopter-illustration}~(j)) that uses an extended Kalman filter (EKF) to produce high-frequency odometry by fusing information from an IMU sensor and a monocular camera. Darivianakis~\etal~\cite{uav-background_quadcopter_darivianakis_icra14} build a quadcopter (Fig.~\ref{fig:drone-quadcopter-illustration}~(k)) that can physically interact with the infrastructures that are being inspected. Driessens and Pounds~\cite{uav-background_quadcopter_driessens_iros13} present a ``Y4'' quadcopter (Fig.~\ref{fig:drone-quadcopter-illustration}~(l)) that combines the simplicity of a conventional quadcopter and the energy efficiency of a helicopter.


\begin{figure*}
  \centering
  \includegraphics[width=\textwidth]{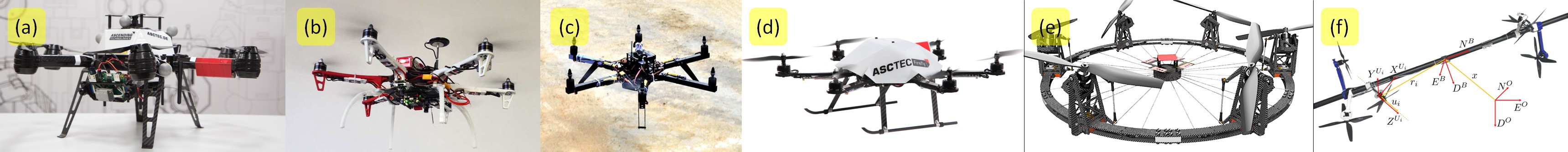}
  \caption{Prototypes of hexacopters (in alphabetical order:~\cite{uav-background_hexacopter_burri_icra16, uav-background_hexacopter_ryll_iros16, 6943040, uav-background_hexacopter_park_iros16, uav-background_hexacopter_zhou_icra14, uav-background_hexacopter_nguyen_iros16}) appear in the reviewed papers. See text for details.}
  \label{fig:drone-hexacopter-illustration}
\end{figure*}

\subsection{Hexacopters}
\label{subsection:state-of-the-art-hexacopters}

A hexacopter (Fig.~\ref{fig:drone-hexacopter-illustration}~(a)--(f))~\cite{uav-background_hexacopter_burri_icra16, uav-background_hexacopter_ryll_iros16, 6943040, uav-background_hexacopter_park_iros16, uav-background_hexacopter_zhou_icra14, uav-background_hexacopter_nguyen_iros16} is a UAV with six rotors. Burri~\etal~\cite{uav-background_hexacopter_burri_icra16} use a hexacopter (Fig.~\ref{fig:drone-hexacopter-illustration}~(a)) to perform system identification study. Specifically, they collect information from an onboard IMU sensor, motor speeds, and hexacopter's pose to estimate the complex dynamic model (required for accurate positioning flight). In addition to the IMU sensor, Zhou~\etal~\cite{uav-background_hexacopter_zhou_icra14} combine visual information from two downward-facing cameras to perform visual odometry in their hexacopter (Fig.~\ref{fig:drone-hexacopter-illustration}~(b)). On the other hand, Yol~\etal~\cite{6943040} demonstrate a hexacopter (Fig.~\ref{fig:drone-hexacopter-illustration}~(c)) that could perform vision-based localization by using a downward-looking camera and geo-referenced images. 
Navigation and obstacle avoidance are also important topics for UAVs. Nguyen~\etal~\cite{uav-background_hexacopter_nguyen_iros16} demonstrate their real-time path planning and obstacle avoidance algorithms with a commercial hexacopter (Fig.~\ref{fig:drone-hexacopter-illustration}~(d)).

Similar to a conventional quadcopter, a conventional hexacopter is a non-holonomic aircraft, which cannot move horizontally without changing its attitude. Ryll~\etal~\cite{uav-background_hexacopter_ryll_iros16} present a hexacopter (Fig.~\ref{fig:drone-hexacopter-illustration}~(e)) that could transform itself from a conventional hexacopter to a holonomic hexacopter, i.e., able to move horizontally without tilting the aircraft, by using a servo to tilt the six rotors simultaneously. Similarly, Park~\etal~\cite{uav-background_hexacopter_park_iros16} design a special hexacopter with six asymmetrically aligned and bi-directional rotors, which enable the hexacopter (Fig.~\ref{fig:drone-hexacopter-illustration}~(f)) to perform fully-actuated flight. While The holonomic capability of a hexacopter is not as energy-efficient as a conventional hexacopter, it has several merits such as robust to wind disturbance, precision flight, and intuitive human-drone interaction. 

\begin{figure*}
  \centering
  \includegraphics[width=\textwidth]{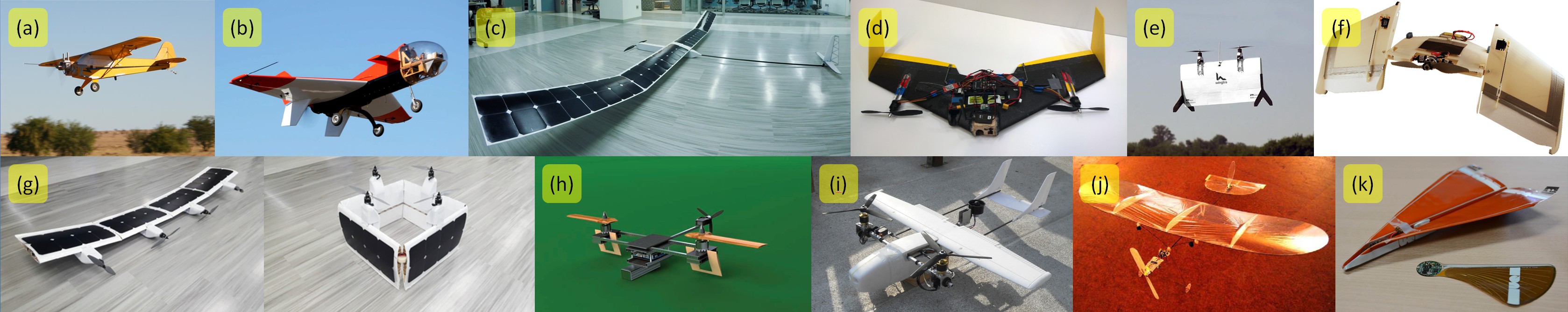}
  \caption{Prototypes of fixed-wing UAVs (in alphabetical order:~\cite{uav-background_fixed-wing_bryson_iros11, uav-background_fixed-wing_hemakumara_tro13, uav-background_fixed-wing_morton_iros15, uav-background_fixed-wing_bapst_iros15, uav-background_fixed-wing_verling_icra16, uav-background_fixed-wing_daler_iros15, uav-background_fixed-wing_sa_iros16, uav-background_fixed-wing_alexis_icra16, uav-background_fixed-wing_papachristos_icra13, uav-background_fixed-wing_zufferey_icra05, uav-background_fixed-wing_pounds_icra13}) appear in the reviewed papers. See text for details.}
  \label{fig:drone-fixedwing-illustration}
\end{figure*}

\subsection{Fixed-wing UAVs}
\label{subsection:state-of-the-art-fixed-wing-uavs}

A fixed-wing UAV (Fig.~\ref{fig:drone-fixedwing-illustration}~(a)--(k))~\cite{uav-background_fixed-wing_bryson_iros11, uav-background_fixed-wing_hemakumara_tro13, uav-background_fixed-wing_morton_iros15, uav-background_fixed-wing_bapst_iros15, uav-background_fixed-wing_verling_icra16, uav-background_fixed-wing_daler_iros15, uav-background_fixed-wing_sa_iros16, uav-background_fixed-wing_alexis_icra16, uav-background_fixed-wing_papachristos_icra13, uav-background_fixed-wing_zufferey_icra05, uav-background_fixed-wing_pounds_icra13}, also known as airplane, aeroplane, or simply a plane, is one of the most common aircrafts in the aviation history. Compared to a multi-rotor aircraft, a fixed-wing UAV generally has higher flight safety (still able to glide for a long time after engines break down in the air) and longer flight time (much more energy-efficient).

Figure~\ref{fig:drone-fixedwing-illustration}~(a)--(c) show three typical fixed-wing UAVs. Bryson and Sukkarieh~\cite{uav-background_fixed-wing_bryson_iros11} demonstrate a mapping application with their fixed-wing UAV by integrating information from an IMU sensor, a GPS sensor, and a downward-facing monocular camera (Fig.~\ref{fig:drone-fixedwing-illustration}~(a)). Hemakumara and Sukkarieh~\cite{uav-background_fixed-wing_hemakumara_tro13} focus on system identification topic and aim to learn the complex dynamic model of their fixed-wing UAV by using Gaussian processes (Fig.~\ref{fig:drone-fixedwing-illustration}~(b)). Morton~\etal~\cite{uav-background_fixed-wing_morton_iros15} focus on hardware development, where they detail the design and developments of their solar-powered and fixed-wing UAV (Fig.~\ref{fig:drone-fixedwing-illustration}~(c)).

Compared to a multi-rotor UAV, a conventional fixed-wing UAV is more energy efficient during cruise flight but does not have the hovering capability, in which a fixed-wing UAV cannot maintains its position in the air and requires more space for take-off and landing. Bapst~\etal~\cite{uav-background_fixed-wing_bapst_iros15} aim to combine the merits of both types of UAVs, where they present their design, modeling, and control of a UAV that could vertically take-off and land (VTOL) like a multi-rotor UAV and perform cruise flight like a fixed-wing UAV (Fig.~\ref{fig:drone-fixedwing-illustration}~(d)). Verling~\etal~\cite{uav-background_fixed-wing_verling_icra16} present another design of this type of hybrid fixed-wing UAV based on a new modeling and controller approach, where they focus on the smooth and autonomous transition between the VTOL mode and cruise mode (Fig.~\ref{fig:drone-fixedwing-illustration}~(e)). 

Researchers have also explored topics on fixed-wing UAVs with transformable shapes. Daler~\etal~\cite{uav-background_fixed-wing_daler_iros15} build a fixed-wing UAV that could fly in the air and walk on the ground by rotating its wings (Fig.~\ref{fig:drone-fixedwing-illustration}~(f)). D'Sa~\etal~\cite{uav-background_fixed-wing_sa_iros16} present a UAV that could fly in a fixed-wing configuration and perform VTOL in a quadcopter configuration (Fig.~\ref{fig:drone-fixedwing-illustration}~(g)).

Alexis and Tzes~\cite{uav-background_fixed-wing_alexis_icra16} present a hybrid UAV, where the UAV could perform cruise flight like a fixed-wing UAV and perform  hovering flight like a bicopter (Fig.~\ref{fig:drone-fixedwing-illustration}~(h)). Their key design lies on the hybrid wings/propellers' structures: in the fixed-wing configuration, the one-blade structures are fixed at the right positions and act as wings; in the bicopter configuration, the one-blade structures rotate and act as propellers. Papachristos~\etal~\cite{uav-background_fixed-wing_papachristos_icra13} develop another type of hybrid UAV, where the UAV could perform cruise flight like a fixed-wing UAV and perform hovering flight like a tricopter (Fig.~\ref{fig:drone-fixedwing-illustration}~(i)).

Several palm-sized fixed-wing UAVs have also been designed by researchers. Zufferey and Floreano~\cite{uav-background_fixed-wing_zufferey_icra05} design a small fixed-wing UAV that has only 30 grams and capable of navigating autonomously at an indoor environment (Fig.~\ref{fig:drone-fixedwing-illustration}~(j)). Despite its small size, the 30-gram fixed-wing UAV can also avoid obstacle during flight by relying on optical flow techinique. Pounds and Singh~\cite{uav-background_fixed-wing_pounds_icra13} present a novel and low-cost fixed-wing UAV by integrating electronics and lift-producing devices onto a paper aeroplane (Fig.~\ref{fig:drone-fixedwing-illustration}~(k)).

\subsection{Flapping-wing UAVs}
\label{subsection:state-of-the-art-flapping-wing-uavs}

\begin{figure*}
  \centering
  \includegraphics[width=\textwidth]{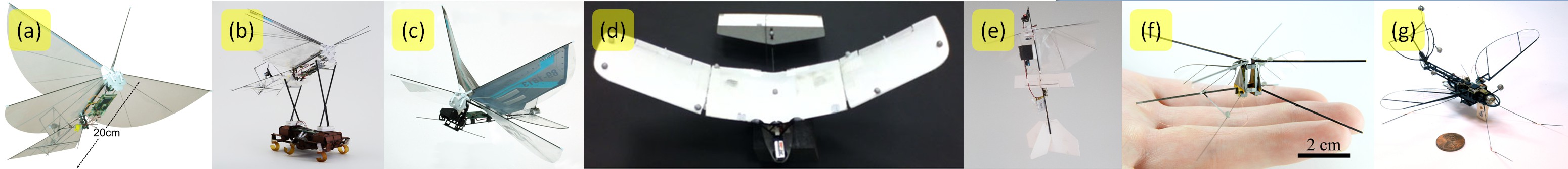}
  \caption{Prototypes of flapping-wing UAVs (in alphabetical order:~\cite{uav-background_flapping-wing_rose_icra14, uav-background_flapping-wing_rose_icra15, uav-background_flapping-wing_peterson_iros11, uav-background_flapping-wing_paranjape_tro13, uav-background_flapping-wing_lamers_iros16, uav-background_flapping-wing_ma_iros15, uav-background_flapping-wing_rosen_icra16}) appear in the reviewed papers. See text for details.}
  \label{fig:drone-flappingwing-illustration}
\end{figure*}

A flapping-wing UAV (Fig.~\ref{fig:drone-flappingwing-illustration}~(a)--(g))~\cite{uav-background_flapping-wing_rose_icra14, uav-background_flapping-wing_rose_icra15, uav-background_flapping-wing_peterson_iros11, uav-background_flapping-wing_paranjape_tro13, uav-background_flapping-wing_lamers_iros16, uav-background_flapping-wing_ma_iros15, uav-background_flapping-wing_rosen_icra16}, also known as ornithopter and usually in about a hand size, is a UAV that generate lifting and forward force by flapping its wings. Aiming for a better aerodynamic modeling, Rose and Fearing~\cite{uav-background_flapping-wing_rose_icra14} compare the flight data collected from a wind tunnel to flight data collected from a free flight condition by using their bird-shaped flapping-wing UAV---H$^{2}$Bird (Fig.~\ref{fig:drone-flappingwing-illustration}~(a)). They find atht the flight data collected from the wind tunnel is not accurate enough to predict the flight data during free flight and further experiments are required. Rose~\etal~\cite{uav-background_flapping-wing_rose_icra15} develop a coordinated launching system for H$^{2}$Bird by mounting it onto a hexapedal robot (Fig.~\ref{fig:drone-flappingwing-illustration}~(b)). With the hexapedal robot's helps, H$^{2}$Bird has a more steady launching velocity. Peterson and Fearing~\cite{uav-background_flapping-wing_peterson_iros11} develop a flapping-wing UAV---BOLT that is capable of flying and walking on the ground like a bipedal robot (Fig.~\ref{fig:drone-flappingwing-illustration}~(c)).

Inspired by birds, Paranjape~\etal~\cite{uav-background_flapping-wing_paranjape_tro13} design a flapping-wing UAV that is able to perch naturally on a chair or human hand (Fig.~\ref{fig:drone-flappingwing-illustration}~(d)). One of the unique features of their UAV is to control the flight path and heading angles by using wing articulation. on the other hand, Lamers~\etal~\cite{uav-background_flapping-wing_lamers_iros16} develop a flapping-wing UAV that has a mini monocular camera system (Fig.~\ref{fig:drone-flappingwing-illustration}~(e)). By combining the camera and a proximity sensor, their UAV can achieve obstacle detection by applying a machine learning method.

Flapping-wing UAVs with insect shapes are also common. Ma~\etal~\cite{uav-background_flapping-wing_ma_iros15} fabricate an bee-shaped flapping-wing UAV that has a mass of 380 mg by using novel methods (Fig.~\ref{fig:drone-flappingwing-illustration}~(f)). After detailing their design and fabrication processes, they also demonstrate a hovering flight with the developed mini UAV. Rosen~\etal~\cite{uav-background_flapping-wing_rosen_icra16} develop another insect-scaled flapping-wing UAV that is capable of flapping and gliding flights (Fig.~\ref{fig:drone-flappingwing-illustration}~(g)).

\subsection{Single-rotor, Coaxial, and Ducted-fan UAVs}
\label{subsection:state-of-the-art-single-rotor-coaxial-and-ducted-fan-uavs}

\begin{figure*}
  \centering
  \includegraphics[width=\textwidth]{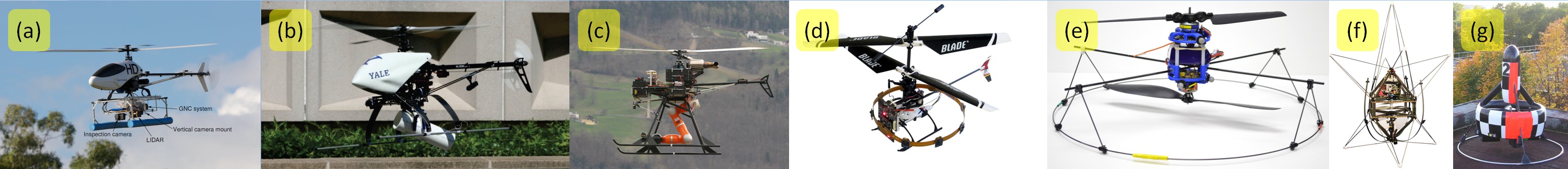}
  \caption{Prototypes of single-rotor helicopters (Fig. (a)--(c))~\cite{uav-background_helicopter_merz_iros11, uav-background_helicopter_backus_iros14, uav-background_helicopter_laiacker_iros16}, coaxial helicopters (Fig. (d)--(f))~\cite{uav-background_coaxial_moore_icra14, uav-background_coaxial_paulos_icra15, uav-background_coaxial_briod_iros13}, and ducted-fan UAV (Fig. (g))~\cite{uav-background_ducted-fan_pflimlin_icra16} appear in the reviewed papers. See text for details.}
  \label{fig:drone-helicopter-illustration}
\end{figure*}

A single-rotor helicopter (Fig.~\ref{fig:drone-helicopter-illustration}~(a)--(c))~\cite{uav-background_helicopter_merz_iros11, uav-background_helicopter_backus_iros14, uav-background_helicopter_laiacker_iros16} is a UAV that relies on a main rotor and a tail rotor to generate thrust for VTOL, hovering, forward, backward, and lateral flights. By using a lidar-based perception system, Merz and Kendoul~\cite{uav-background_coaxial_moore_icra14} demonstrate a helicopter that can perform obstacle avoidance and close-range infrastructure inspection. Backus~\etal~\cite{uav-background_coaxial_paulos_icra15} focus on the aerial manipulation of an helicopter. Specifically, they design a robotic hand for their helicopter to perform grasping and perching actions effectively. Laiacker~\etal~\cite{uav-background_helicopter_laiacker_iros16} aim to optimize their visual servoing system on a helicopter that is equipped with a 7 DoF industrial manipulator.

On the other hand, a coaxial helicopter (Fig.~\ref{fig:drone-helicopter-illustration}~(d)--(f)) is a UAV that uses two contra-rotating rotors mounted on the same axis to generate thrust for VTOL, hovering, forward, backward, and lateral flights. Moore~\etal~\cite{uav-background_coaxial_moore_icra14} implement a lightweight omnidirectional vision sensor for their mini coaxial helicopter to perform visual navigation. Conventionally, a helicopter requires additional servo motor and a mechanical device called swashplate for horizontal position control. Paulos and Yim~\cite{uav-background_coaxial_paulos_icra15} present a novel coaxial helicopter that requires no servo motor and mechanical device for horizontal position control. In order to perform horizontal movement, the rotors are driven by a modulated signal in order to generate both lifting and lateral forces simultaneously. Instead of avoid obstacles like a conventional UAV, Briod~\etal~\cite{uav-background_coaxial_briod_iros13} design a coaxial helicopter that uses force sensors around the UAV to detect obstacle and able to perform autonomous navigation safely without the high risks of collisions. 

A ducted-fan UAV (Fig.~\ref{fig:drone-helicopter-illustration}~(g)) is a UAV that has similar rotors configuration with a coaxial helicopter but the rotors are mounted within a cylindrical duct. The duct helps to reduce thrust losses of the propellers and the ducted fans normally have rotational speeds. Pflimlin~\etal~\cite{uav-background_ducted-fan_pflimlin_icra16} present a ducted-fan UAV that can stabilize itself in wind gusts by using a two-level controller for position and attitude controls.

\subsection{Octocopter, Glider, Blimp, and Ionic Flyer UAVs}
\label{subsection:state-of-the-art-octocopter-glider-blimp-and-ionic-flyer-uavs}

\begin{figure*}
  \centering
  \includegraphics[width=\textwidth]{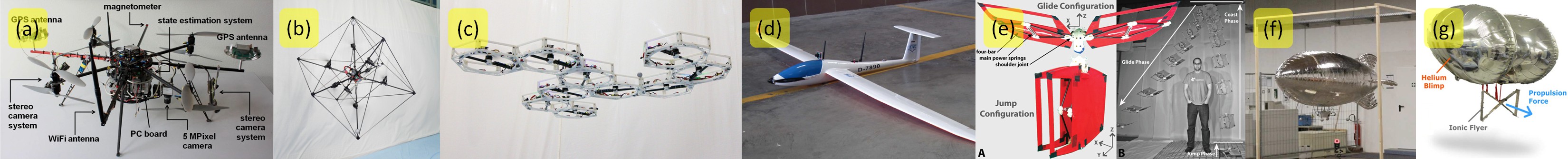}
  \caption{Prototypes of octocopters (Fig. (a)--(b))~\cite{uav-background_octocopter_schneider_icra16, uav-background_octocopter_brescianini_icra16}, multirotor (Fig. (c))~\cite{uav-background_multirotor_oung_ijrr13}, gliders (Fig. (d)--(e))~\cite{uav-background_glider_woodward_ijrr14, uav-background_glider_cobano_iros13}, blimp (Fig. (f))~\cite{uav-background_blimp_muller_iros11}, and Ionic Flyer (Fig. (g))~\cite{uav-background_ionic-flyer_poon_icra16} appear in the reviewed papers. See text for details.}
  \label{fig:drone-gliderblimp-illustration}
\end{figure*}

An octocopter (Fig.~\ref{fig:drone-gliderblimp-illustration}~(a)--(b))~\cite{uav-background_octocopter_schneider_icra16, uav-background_octocopter_brescianini_icra16} is a UAV with eight rotors. Schneider~\etal~\cite{uav-background_octocopter_schneider_icra16} demonstrate an octocopter with a multiple fisheye-camera system that can perform a simultaneous localization and mapping (SLAM) function (Fig.~\ref{fig:drone-gliderblimp-illustration}~(a)). Different from a conventional octocopter, Brescianini and D'Andrea~\cite{uav-background_octocopter_brescianini_icra16} build a octocopter that has eight rotors facing to eight different direction in the 3D space (Fig.~\ref{fig:drone-gliderblimp-illustration}~(b)). This unique configuration allows the UAV to have 6 DoF and to hover stably at any attitude. More importantly, the octocopter is able to control its force in the 3D space and is useful for applications such as aerial manipulation. 

A multirotor is a UAV with more than one rotor and has simple rotors configuration for flight control. Oung and D'Andrea~\cite{uav-background_multirotor_oung_ijrr13} design a modular multirotor system, where each rotor aircraft has a hexagonal shape and can be assembled into a multirotor aircraft with different configuration. With a distributed state estimation algorithm and a parameterized control strategy, the multirotor is able to fly in any flight-feasible configuration both indoors and outdoors (Fig.~\ref{fig:drone-gliderblimp-illustration}~(c)).

A glider (Fig.~\ref{fig:drone-gliderblimp-illustration}~(d)--(e))~\cite{uav-background_glider_cobano_iros13, uav-background_glider_woodward_ijrr14} is a UAV that uses its wings and aerodynamics to glide in the air. Glider normally has a outlook like a fixed-wing UAV but does not rely on an active propulsion system during gliding performance. For instance, Cobano~\etal~\cite{uav-background_glider_cobano_iros13} demonstrate multiple gliders that can glide cooperatively in the sky (Fig.~\ref{fig:drone-gliderblimp-illustration}~(d)). To glide for a long time in the sky without active propulsion control, the gliders detect thermal currents and exploit their energy to soar and continue to glide in the air. Inspired by a vampire bat, Woodward and Sitti~\cite{uav-background_glider_woodward_ijrr14} build a different type of glider UAV, where their UAV can jump from the ground and then uses its wings to glide in the air (Fig.~\ref{fig:drone-gliderblimp-illustration}~(e)).

A blimp, also known as a non-rigid airship, is a lighter-than-air UAV that relies on helium gas inside an envelope to generate lifting force. Different from a Montgolfi{\`e}re or hot air balloon, a blimp keeps its envelope shape with the internal pressure of helium gas and has actuation units for motion control. By using a motion capture system, M{\"u}ller and Burgard~\cite{uav-background_blimp_muller_iros11} present an autonomous blimp that can navigate in an indoor environment with an online motion planning method (Fig.~\ref{fig:drone-gliderblimp-illustration}~(f)). Poon~\etal~\cite{uav-background_ionic-flyer_poon_icra16} aim to design a UAV that is noiseless and vibration-free by using ionic propulsion, where they call their UAV Inoic Flyer (Fig.~\ref{fig:drone-gliderblimp-illustration}~(g)). Instead of using rotors, they create a propulsion unit that has no moving mechanic parts and relies on high electrical voltage to create thrust by accelerating ions.

\subsection{Cyclocopter, Spincopter, Coand{\v a}, Parafoil, and Kite UAVs}
\label{subsection:state-of-the-art-cyclocopter-spincopter-coanda-parafoil-and-kite-uavs}

\begin{figure*}
  \centering
  \includegraphics[width=\textwidth]{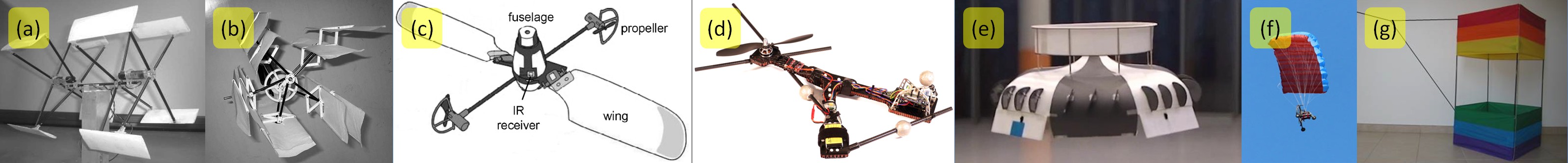}
  \caption{Prototypes of cyclocopters (Fig. (a)--(b))~\cite{uav-background_rotary-wing_tanaka_tom07, uav-background_rotary-wing_hara_tro09}, spincopters (Fig. (c)--(d))~\cite{uav-background_spinning-wing_orsag_icra11, uav-background_spinning-wing_zhang_icra16}, Coand{\v a} UAV (Fig. (e))~\cite{uav-background_coanda_han_icra14}, parafoil UAV (Fig. (f))~\cite{uav-background_parafoil_cacan_tom15}, and kite UAV (Fig. (g))~\cite{uav-background_kite_christoforou_icra10} appear in the reviewed papers. See text for details.}
  \label{fig:drone-special-illustration}
\end{figure*}

A cyclocopter (Fig.~\ref{fig:drone-special-illustration}~(a)--(b))~\cite{uav-background_rotary-wing_tanaka_tom07, uav-background_rotary-wing_hara_tro09} is a UAV that flies by rotating a cyclogyro wing with several wings positioned around the edge of a cylindrical structure. Generally, the wings' angles of attack are adjusted collectively by a servo motor to generate required forces. Tanaka~\etal~\cite{uav-background_rotary-wing_tanaka_tom07} built a cyclocopter that can the angles of attack using a novel eccentric point mechanism without additional actuators (Fig.~\ref{fig:drone-special-illustration}~(a)). Hara~\etal~\cite{uav-background_rotary-wing_hara_tro09} developed a cyclocopter based on a pantograph structure, where diameters of the wings can be expanded or contracted (with reference to the rotational axis) for flight control (Fig.~\ref{fig:drone-special-illustration}~(b)).

A spincopter (Fig.~\ref{fig:drone-special-illustration}~(c)--(d))~\cite{uav-background_spinning-wing_orsag_icra11, uav-background_spinning-wing_zhang_icra16} is a UAV that spins itself during flight. Orsag~\etal~\cite{uav-background_spinning-wing_orsag_icra11} designed a spincopter that can spin the central wings (and the whole aircraft) using two small motors mounted at the edge of the virtual ring of the UAV (Fig.~\ref{fig:drone-special-illustration}~(c)). The motors adjust their output thrust symmetrically/asymmetrically for vertical/horizontal motion control. By using a asymmetrical design and cascaded control strategy, Zhang~\etal~\cite{uav-background_spinning-wing_zhang_icra16} demonstrate a spincopter that has three translational DoF and two rotational DoF with only one rotor (Fig.~\ref{fig:drone-special-illustration}~(d)).

A Coand{\v a} UAV is an aircraft that produces lifting force by utilizing the Coand{\v a} effect.  Specifically, the Coand{\v a} effect is caused by the tendency of a jet of fluid  to follow an adjacent surface and to attract the surrounding fluid. Thanks to the Bernoulli principle, in which pressure is low when speed is high, a Coand{\v a} UAV can generate enough lifting force to hover in the air when the Coand{\v a} effect is strong enough. Han~\etal~\cite{uav-background_coanda_han_icra14} developed a Coand{\v a} UAV in a flying saucer shape (Fig.~\ref{fig:drone-special-illustration}~(e)). By attaching additional servo motors onto the UAV for flap control, their Coand{\v a} UAV is able to perform VTOL and horizontal movements in the air.

Parafoil UAVs (Fig.~\ref{fig:drone-special-illustration}~(f))~\cite{uav-background_parafoil_cacan_tom15} and kite UAVs (Fig.~\ref{fig:drone-special-illustration}~(g))~\cite{uav-background_kite_christoforou_icra10} resemble the shapes and flying principles of a parafoil or kite. For an aerial cargo delivery application, Cacan~\etal~\cite{uav-background_parafoil_cacan_tom15} improved the landing accuracy of an autonomous parafoil UAV with the assistance of a ground-based wind measurement system (Fig.~\ref{fig:drone-special-illustration}~(f)) while Christoforou develop a robotic kite UAV that can surf automatically in the air (Fig.~\ref{fig:drone-special-illustration}~(g))~\cite{uav-background_kite_christoforou_icra10}.

\section{Discussion and Final Remarks}
\label{sec:discussion-and-final-remarks}


While we have covered and selected more than one thousand UAV papers in several top journals and conferences since 2001, we focused on the robotic communities (TRO, TME, IJRR, RAS, IROS, ICRA, HRI, ROMAN). To extend the survey, we could expand into journals/conferences in the aerospace and aeronautics communities, such as International Journal of Robust and Nonlinear Control~\cite{conference_ijrnc}, Journal of Guidance, Control, and Dynamics~\cite{conference_jgcd}, and International Conference on Unmanned Aircraft Systems~\cite{conference_icuas}.  
UAVs from the commercial sectors would also provide fertile ground. To name a few, the DJI Phantom 4~\cite{uav_product_dji-phantom-4} and YUNEEC Typhoon H~\cite{uav_product_yuneec-typhoon-h}) appear to possess advanced path planning, human tracking, and obstacle avoidance algorithms. However, private companies often do not provide detailed technical information to the public.


In the Part II of this survey, we cover the trends in aerial robotics by discussing three emerging topics---(i) holonomic UAVs, a special type of UAV that can perform horizontal motions while maintaining orientation; (ii) localization and mapping with UAV; and (iii) human-drone interaction.



\appendices

\section{Additional Survey Results}
\label{apx:additional_survey_results}

\begin{figure}
  \centering
  \includegraphics[trim={0.5cm -2cm -0.5cm 0cm},clip,width=0.48\textwidth]{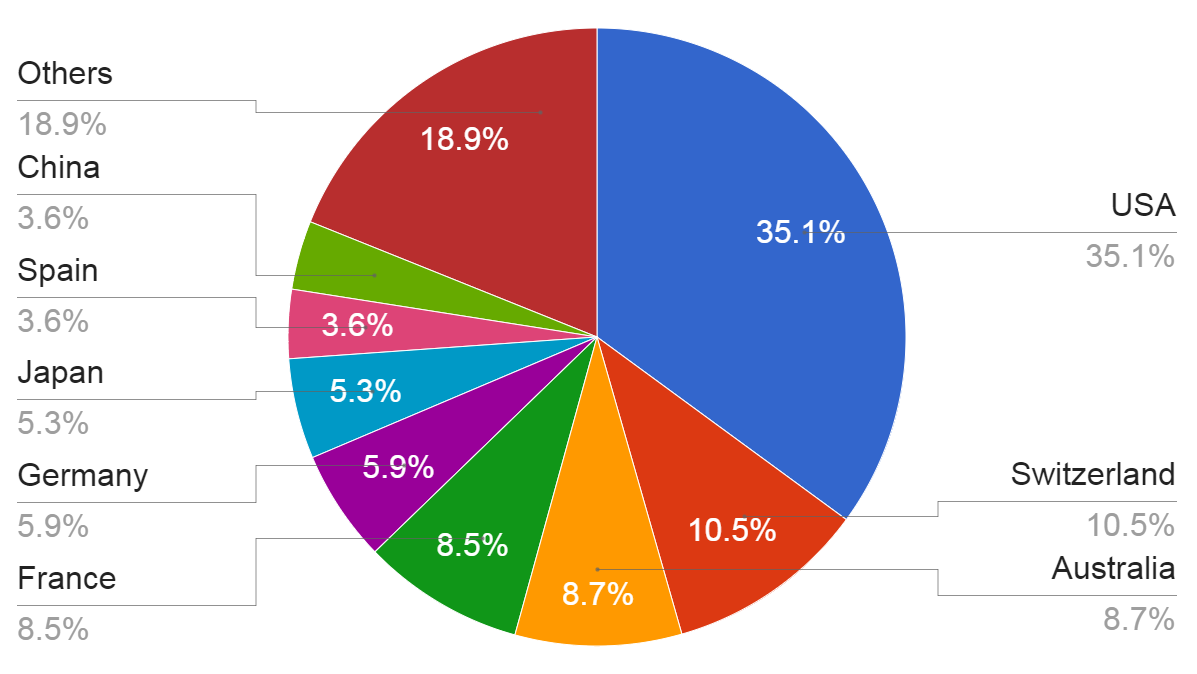}
  \caption{Pie chart of the country distribution of UAV papers. Best viewed in color. See text for details.}
  \label{fig:papers-distribution-by-country}
\end{figure}

\begin{figure}
  \centering
  \includegraphics[trim={0.5cm 0cm -0.5cm 0cm},clip,width=0.48\textwidth]{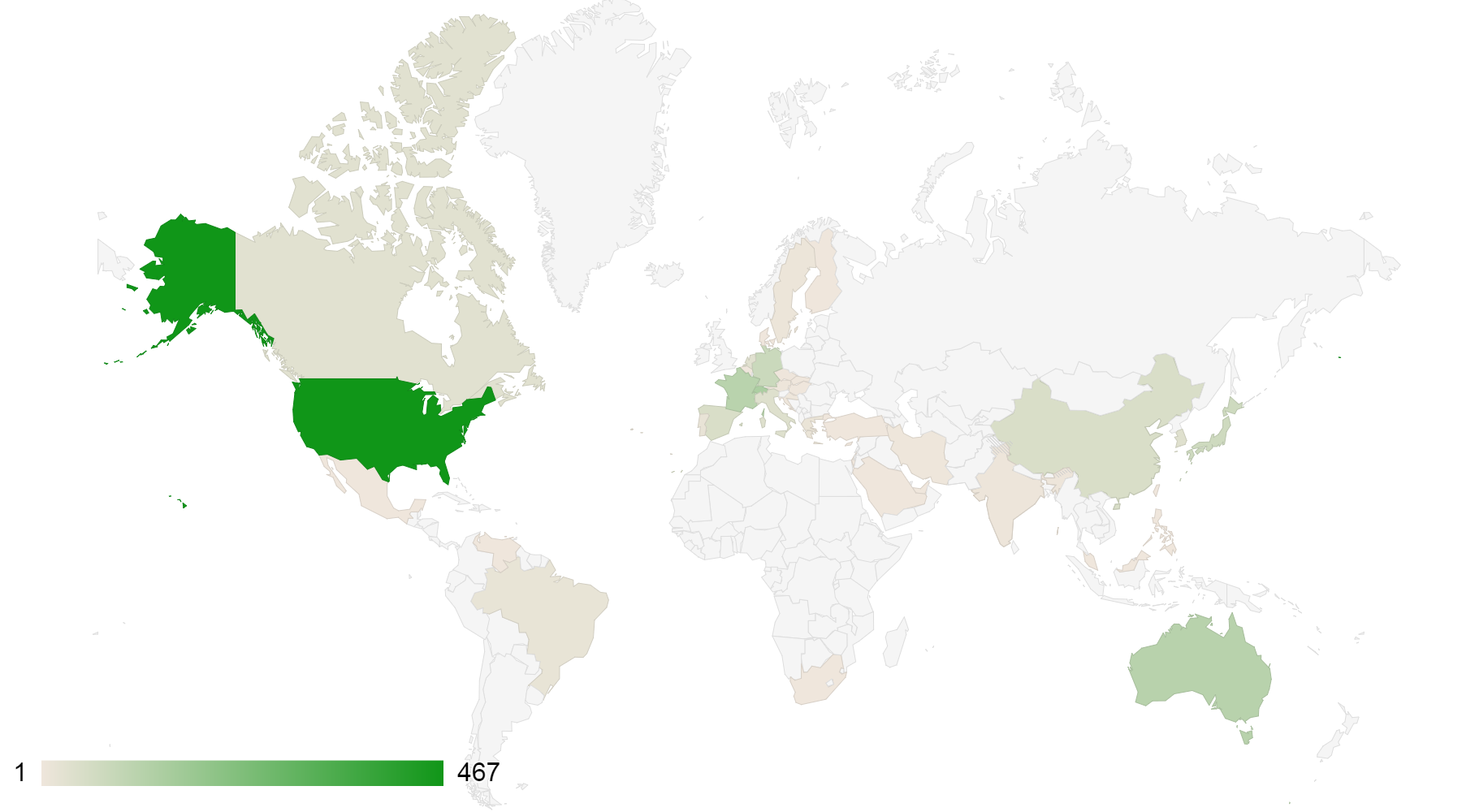}
  \caption{Map of the country distribution of UAV papers. Best viewed in color. See text for details.}
  \label{fig:papers-distribution-by-country-2}
\end{figure}

Figure~\ref{fig:papers-distribution-by-country} and Fig.~\ref{fig:papers-distribution-by-country-2} show a pie chart and a map of the country distribution of UAV papers. In descending order, the top ten countries with the most drone papers since 2001 are United States of America, Switzerland, France, Australia, Germany, Japan, Spain, China, Italy, and South Korea. Other countries in the pie chart include Canada, Mexico, Venezuela, Brazil, United Kingdom, Portugal, Netherlands, Belgium, Austria, Czech Republic, Croatia, Hungary, Slovakia, Sweden, Finland, Denmark, Greece, Cyprus, South Africa, United Arab Emirates, Iran, Israel, Saudi Arabia, Turkey, India, Taiwan, Philippines, Malaysia, and Singapore.

\begin{figure}
  \centering
  \includegraphics[trim={0.5cm 0cm -0.5cm 0cm},clip,width=0.48\textwidth]{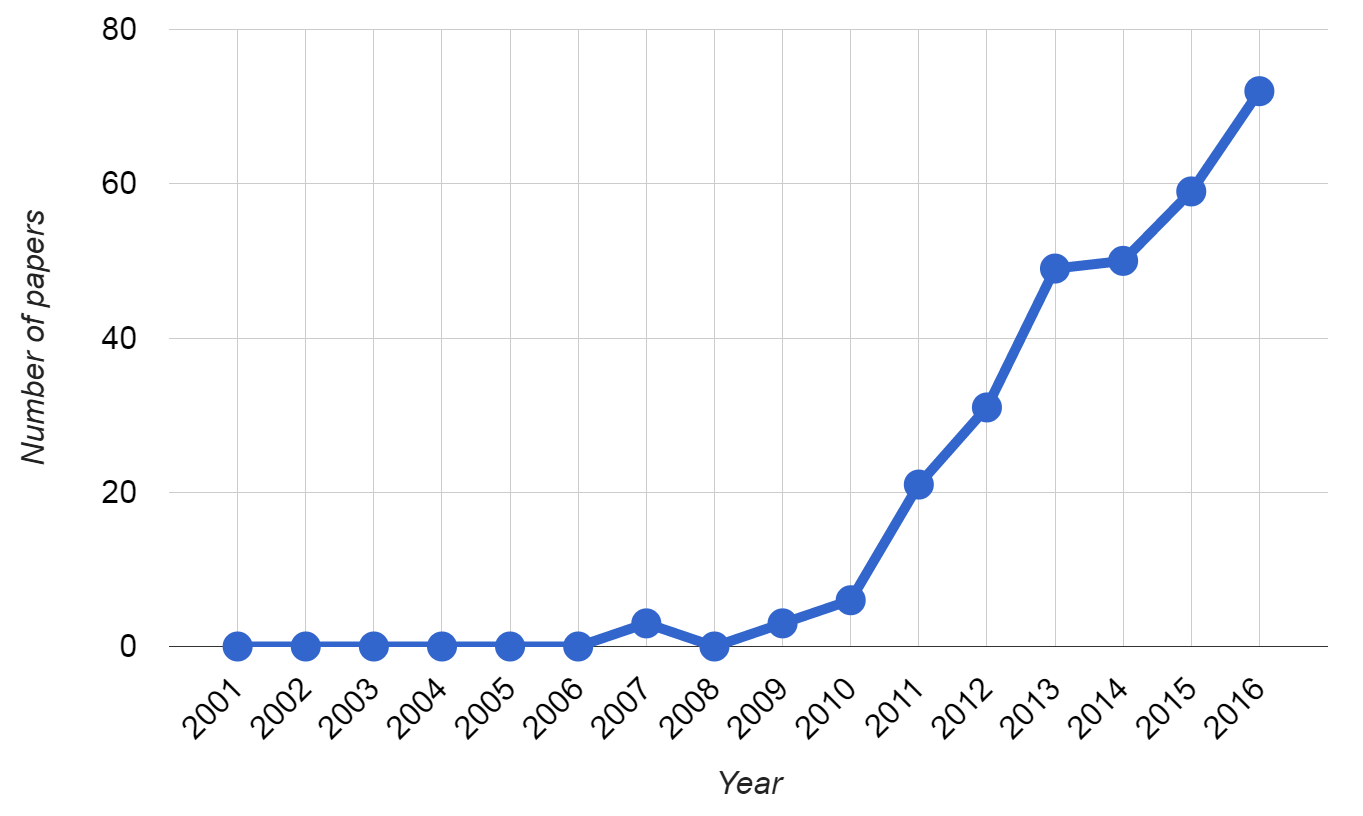}
  \caption{Number of UAV papers that use motion capture system during the flight experiments over the years.}
  \label{fig:papers-with-mocap-system}
\end{figure}

Figure~\ref{fig:papers-with-mocap-system} shows that the number of papers that use motion capture system increases rapidly since 2009. Motion capture system is a system that relies on multiple high-speed cameras to record the movement of reflective markers attached on the UAV in real time. The system offers sub-millimeter position accuracy measurement and is useful in various drone researches, such as topics related to positioning control and visual localization. In term of occurrence (reputation) in descending order, the top widely used commercial system are Vicon~\cite{mocap_system_vicon}, OptiTrack~\cite{mocap_system_opti-track}, Qualisys~\cite{mocap_system_qualisys}, MotionAnalysis~\cite{mocap_system_motion-analysis}, PTI Phoenix~\cite{mocap_system_pti-phoenix}, Advanced Realtime Tracking (ART)~\cite{mocap_system_advanced-realtime-tracking}, NaturalPoint~\cite{mocap_system_natural-point}, and Leica~\cite{mocap_system_leica}.

\ifCLASSOPTIONcaptionsoff
  \newpage
\fi


\bibliographystyle{IEEEtran}
\bibliography{IEEEabrv,IEEEexample,bib_takeishi}

%

\begin{IEEEbiography}[{\includegraphics[width=1in,height=1.25in,clip,keepaspectratio]{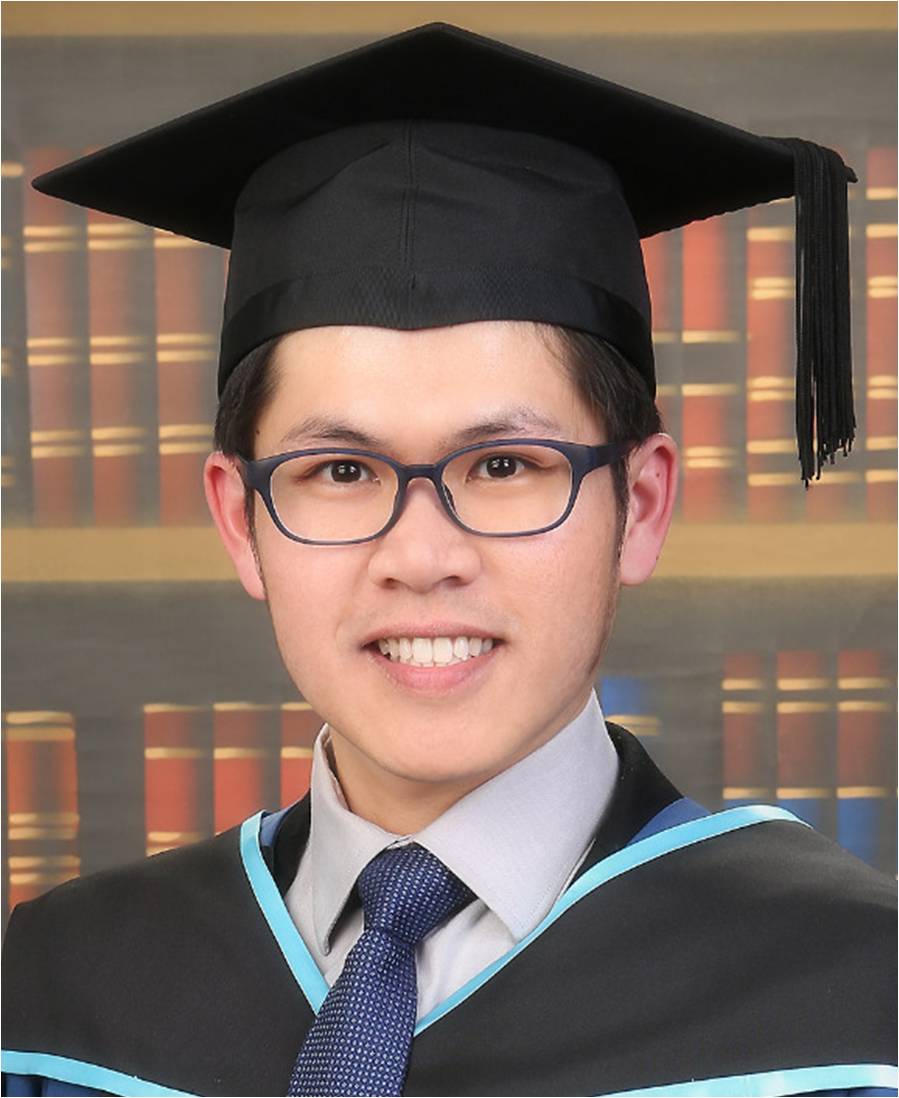}}]{Chun Fui Liew}
received his B.Eng. degree from the Nanyang Technological University, Singapore and M.Sc. degree from the National University of Singapore. Recently, he received his M.Eng. and Ph.D. degrees in Aeronautics and Astronautics Engineering from the University of Tokyo, Japan. He is now a drone researcher and developer with the Hongo Aerospace Inc. His research interests include aerial robotics, pattern recognition, machine learning, and computer vision.
\end{IEEEbiography}

\begin{IEEEbiography}
[{\includegraphics[width=1in,height=1.25in,clip,keepaspectratio]{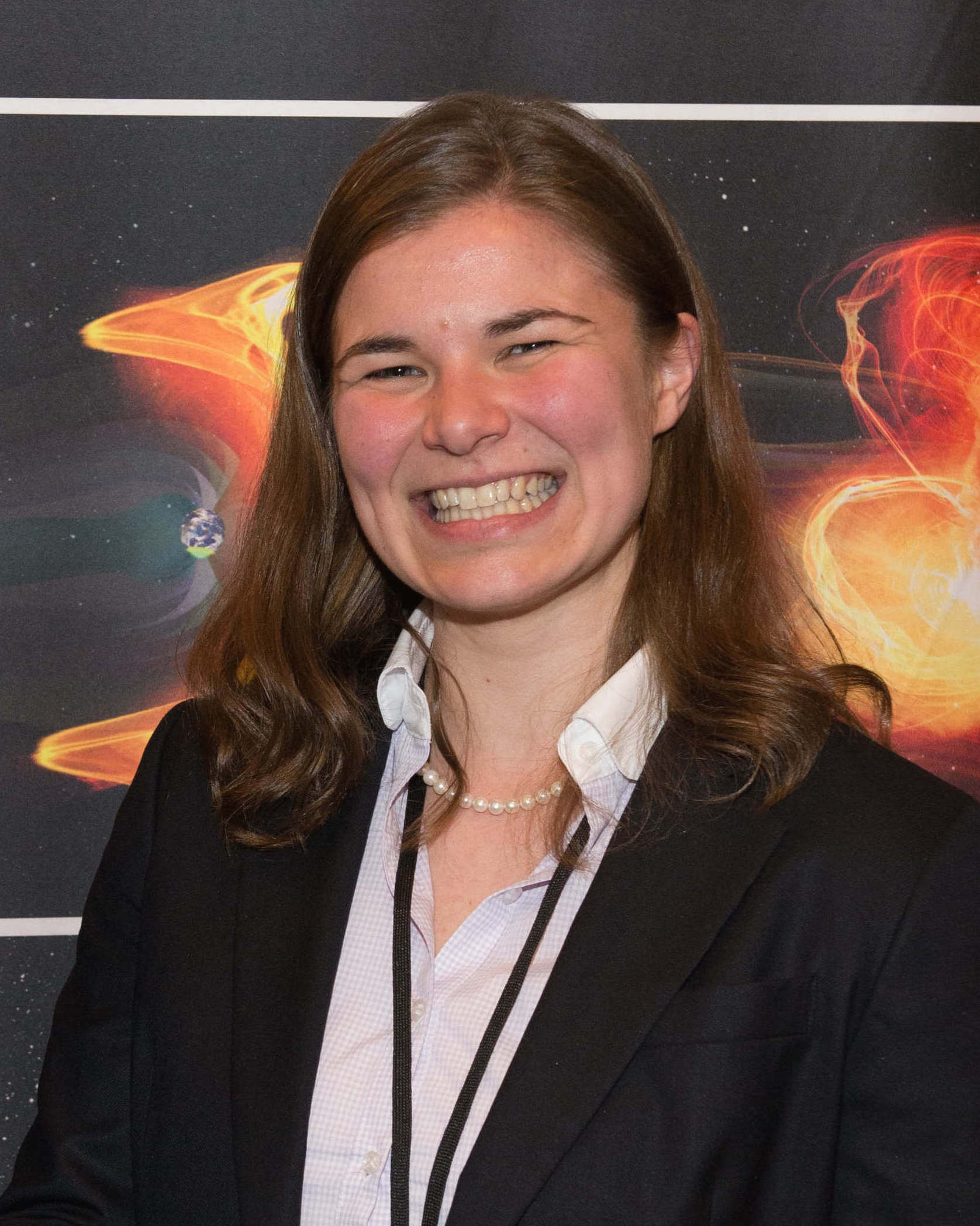}}]{Danielle DeLatte}
is an aerospace engineer with passions for human-robot interaction and outreach. She is currently a Ph.D. student at the University of Tokyo and earned her S.B. Aerospace Engineering with Information Technology from MIT and M.Sc. in Space Studies with Human Space Flight at the International Space University. Prior to her Ph.D. studies, she worked on space robotics at NASA. She is a member of the American Institute of Aeronautics and Astronautics (AIAA) and the Institute of Electrical and Electronics Engineers (IEEE). 
\end{IEEEbiography}

\begin{IEEEbiography}[{\includegraphics[width=1in,height=1.25in,clip,keepaspectratio]{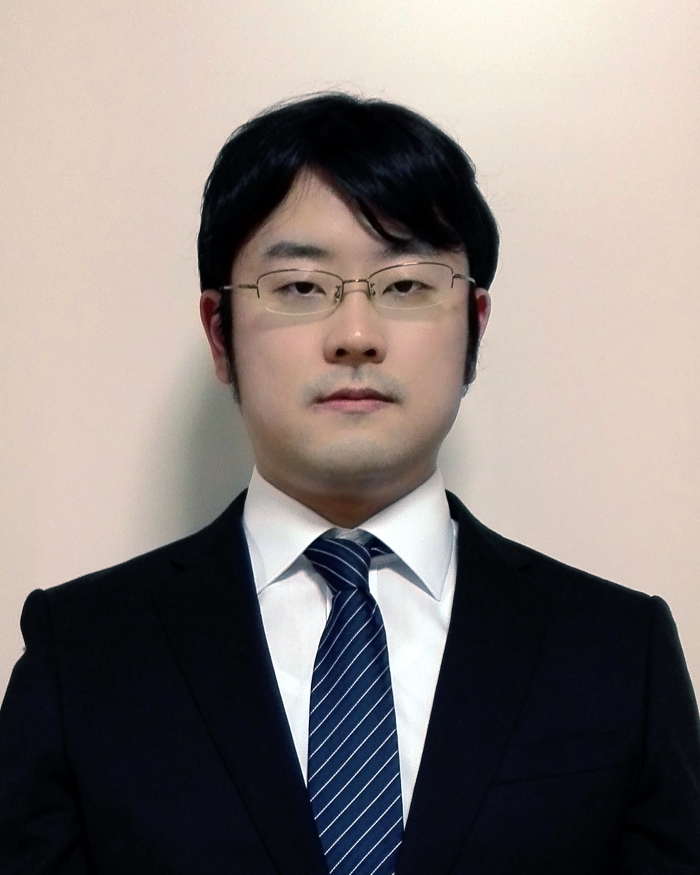}}]{Naoya Takeishi}
received his M.Eng. degree in Aeronautics and Astronautics Engineering from the University of Tokyo, Japan. He is currently a Ph.D. student at the University of Tokyo, Japan. His research interest include machine learning, analysis of nonlinear dynamical systems and intelligent robotics. He is a member of the Japanese Society for Artificial Intelligence (JSAI) and the Japan Society for Aeronautical and Space Sciences (JSASS).
\end{IEEEbiography}


\begin{IEEEbiography}
[{\includegraphics[width=1in,height=1.25in,clip,keepaspectratio]{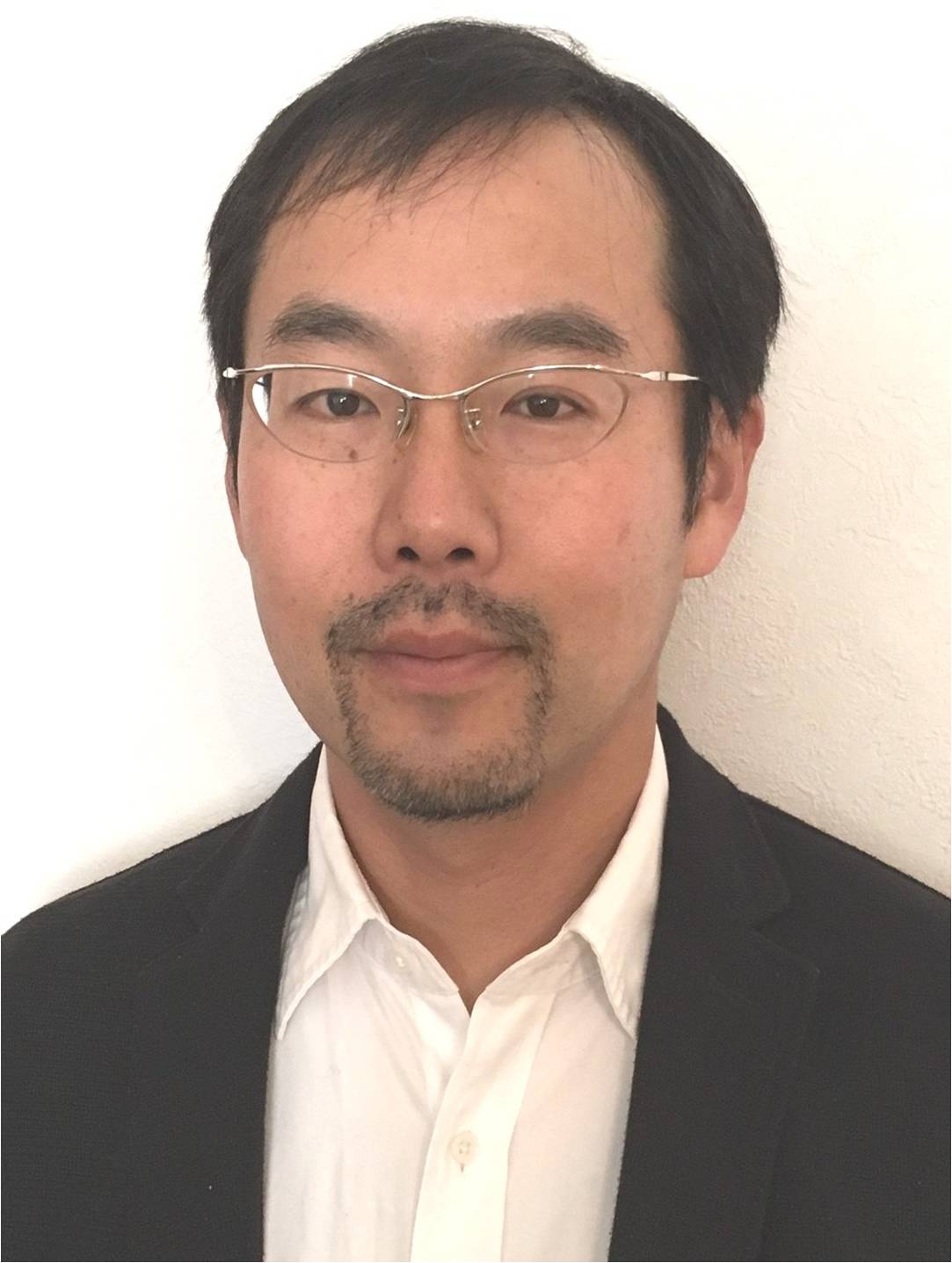}}]{Takehisa Yairi}
received his M.Eng. and Ph.D. degrees from the University of Tokyo, Japan in 1996 and 1999 respectively. He is currently a full-time Associate Professor with the Graduate School of Engineering in the University of Tokyo. His research interests include data mining, machine learning, mobile and space robotics. He is a member of The Japanese Society for Artificial Intelligence (JSAI) and The Robotics Society of Japan (RSJ).
\end{IEEEbiography}




\end{document}